\PassOptionsToPackage{table}{xcolor}
\documentclass{article} 
\usepackage[final]{colm2026_conference}

\usepackage{microtype}
\usepackage{subcaption}
\usepackage{booktabs} 
\usepackage{caption}
\usepackage{adjustbox}
\usepackage{hyperref}
\usepackage{xcolor}
\usepackage{soul}

\usepackage{booktabs}
\usepackage{multirow}
\usepackage{graphicx}
\usepackage{algorithm}
\usepackage{algpseudocode}
\usepackage{amsmath}
\usepackage{amssymb}
\usepackage{mathtools}
\usepackage{amsthm}
\usepackage{enumitem}
\usepackage{float}
\usepackage{xcolor}
\usepackage{caption}
\usepackage{tcolorbox}
\usepackage{tikz}
\usepackage{wrapfig}
\usepackage[capitalize, noabbrev]{cleveref}

\usepackage{placeins}

\theoremstyle{plain}

\theoremstyle{definition}

\theoremstyle{remark}

\usepackage[textsize=tiny]{todonotes}
\usepackage{url}
\usepackage{booktabs}
\usepackage{fontawesome5}
\newcommand{\projecticon}{%
  \raisebox{-0.15ex}{\includegraphics[height=1em]{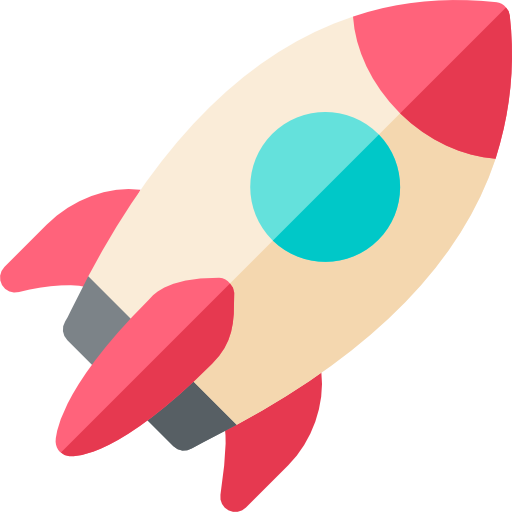}}%
}
\newcommand{\hficon}{%
  \raisebox{-0.15ex}{\includegraphics[height=0.85em]{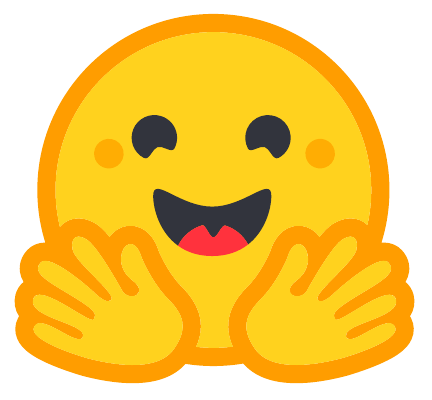}}%
}

\usepackage{lineno}
\usepackage{xcolor}
\usepackage{colortbl}

\definecolor{bestgreen}{RGB}{198, 239, 206}
\definecolor{darkblue}{rgb}{0, 0, 0.5}
\hypersetup{
    colorlinks=true,
    citecolor=darkblue,
    linkcolor=darkblue,
    urlcolor=darkblue
}

\title{Simulating Organized Group Behavior:\\ New Framework, Benchmark, and Analysis}

\author{Xinkai Zou\textsuperscript{1}\thanks{Equal contribution.}, 
Yiming Huang\textsuperscript{1}\footnotemark[1], 
Zhuohang Wu\textsuperscript{2}, 
Jian Sha\textsuperscript{1}, 
Nan Huang\textsuperscript{1}, \\
\textbf{Longfei Yun}\textsuperscript{3}, 
\textbf{Jingbo Shang}\textsuperscript{1}\thanks{Corresponding author.},
\textbf{Letian Peng}\textsuperscript{1}\footnotemark[2]\\
\textsuperscript{1}University of California, San Diego \quad 
\textsuperscript{2}University of California, Irvine \quad
\textsuperscript{3}Meta \\
\texttt{\{x9zou, yih112, n5huang, loyun, jshang, lepeng\}@ucsd.edu} \\
\texttt{zhuohaw2@uci.edu, bekissey@gmail.com}\\[0.5em]
\small
\projecticon\ \href{https://jayzou3773.github.io/projects/group-behavior-simulation/}{Project Page}
\quad
\faGithub\ \href{https://github.com/jayzou3773/Organized-Group-Behavior-Simulation}{Code Repo}
\quad
\hficon\ \href{https://huggingface.co/datasets/jayzou3773/GROVE}{Benchmark}
}

\definecolor{glm}{HTML}{AA66DD}

\begin{document}
    \ifcolmsubmission \linenumbers \fi

    \maketitle

    \begin{abstract}
    Simulating how organized groups (e.g., corporations) make decisions (e.g., responding to a competitor's move) is essential for understanding real-world dynamics and could benefit relevant applications (e.g., market prediction).
    In this paper, we formalize this problem as a concrete research platform for group behavior understanding, providing: (1) a task formalization with benchmark and evaluation criteria, (2) a structured and adaptive analytical framework, and (3) detailed temporal and cross-group analysis.
    Specifically, we propose \textbf{Organized Group Behavior Simulation}, a task that models organized groups as collective entities from a practical perspective: given a group facing a particular situation (e.g., AI Boom), predict the decision it would take. To support this task, we present \textbf{GROVE} (\textbf{GR}oup \textbf{O}rganizational Beha\textbf{V}ior \textbf{E}valuation), a benchmark covering 44 organized groups with 8,052 real-world context--decision pairs collected from Wikipedia and TechCrunch\footnote{TechCrunch is an American global online newspaper focusing on topics regarding high-tech and startup companies. \url{https://techcrunch.com/}} across 9 domains, with an end-to-end evaluation protocol assessing consistency, initiative, scope, magnitude, and horizon.
    Beyond straightforward prompting pipelines, we propose a structured analytical framework that converts collective decision-making events into an interpretable, adaptive, and traceable behavioral model, achieving stronger performance than summarization- and retrieval-based baselines. 
    It further introduces an adapter mechanism for time-aware evolution and group-aware transfer, and traceable evidence nodes grounding each decision rule in originating historical events. Our analysis reveals temporal behavioral drift within individual groups, which the time-aware adapter effectively captures for stronger prediction, and structured cross-group similarity that enables knowledge transfer for data-scarce organizations.
    \end{abstract}

    \section{Introduction}


Organized groups, guided by institutional routines and collective norms, exhibit structured behavioral patterns that can be systematically modeled~\citep{cyert1963behavioral}. Simulating how these organized groups make decisions is key to understanding real-world dynamics and could benefit applications such as market forecasting, policy analysis, and scenario planning~\citep{march1958organizations, north1990institutions}. For instance, predicting how a corporation would respond to a competitor's product launch, or how a nation would act in the face of a territorial incursion. 
Recently, large language models (LLMs) have been increasingly adopted for individual behavior simulation~\citep{park2023generative, horton2023large, bobrov2025fomc, baker2024senate}, yet group behavior simulation still lacks an established framework. This motivates several critical questions: (i) Can we formalize the task of simulating organized group behavior and systematically benchmark it? (ii) Can we design methods that surpass naive prompting to provide trustworthy model of group behavior? (iii) Can such a system track the temporal dynamics and enable robust cross-group~comparisons?

\begin{figure}[t] 
    \centering
    \includegraphics[width=\linewidth]{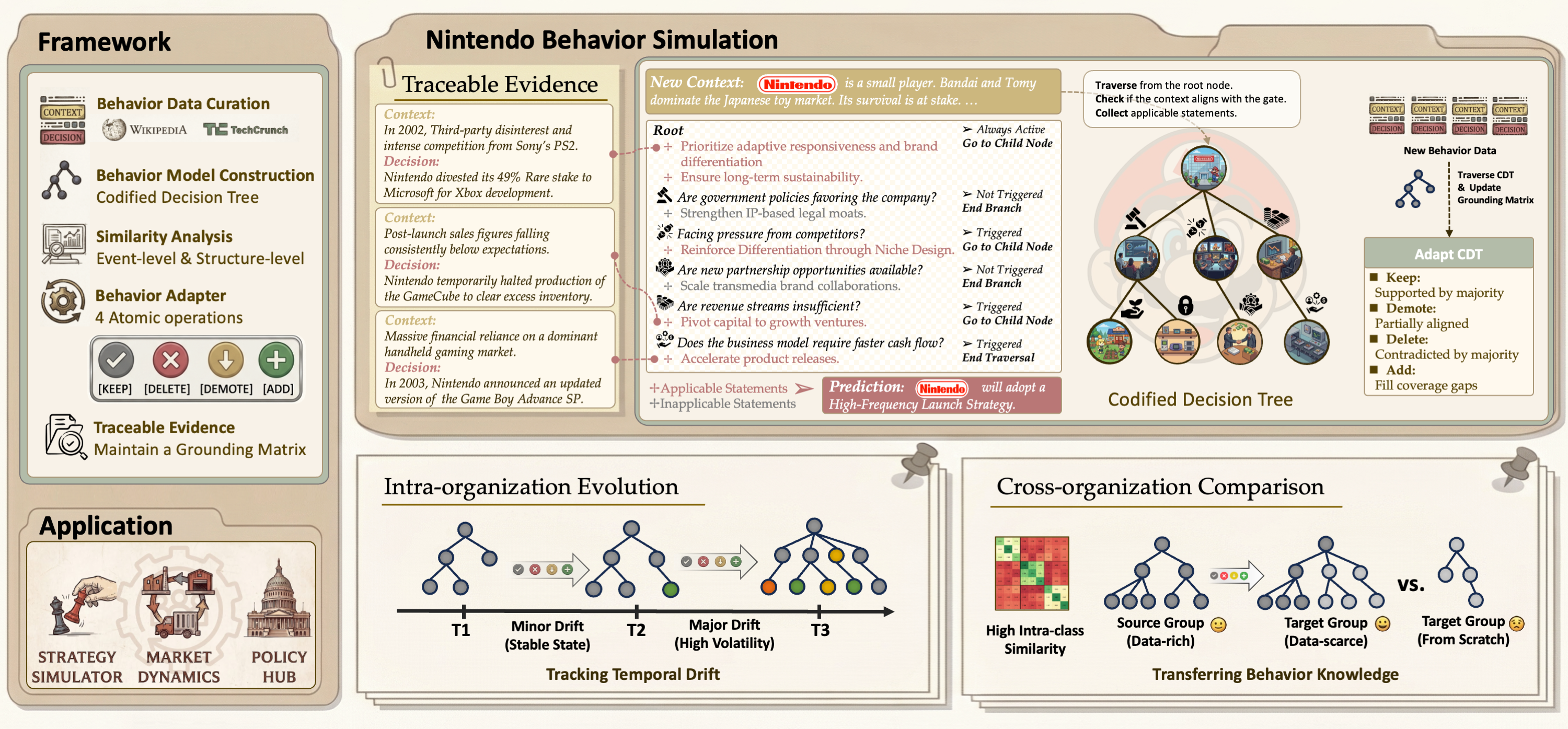} 
    \caption{Framework for Simulating Organized Group Behavior.
    }
    \label{fig:main}
\end{figure}


We formalize \textbf{Organized Group Behavior Simulation} as predicting the decisions an organized group would make when facing a new situation, based on its historical context-decision records. As shown in Figure~\ref{fig:main}, each context captures the situational background (e.g., market shifts), and each decision records the group's concrete response (e.g., product launches). 
For benchmarking, we introduce \textbf{GROVE} (\textbf{GR}oup \textbf{O}rganizational Beha\textbf{V}ior \textbf{E}valuation), to evaluate group behavior simulation with 8,052 real-world context--decision pairs. The dataset is constructed through an automatic pipeline applied to two complementary sources. The Wikipedia part spans 35 organized groups across 9 domains, with records covering decades to over a century. The TechCrunch part focuses on 9 prominent technology corporations, providing up-to-date decision records from Jan. 2025 to Mar. 2026, which postdate most current LLMs' training cutoffs. To ensure comprehensive evaluation, GROVE employs a multi-dimensional criterion that assesses predictions across 5 key dimensions: consistency, initiative, scope, magnitude, and horizon. 


Real-world context--decision data spans extended time horizons triggering temporal behavior drift and covers diverse organizational contexts for cross-group analysis, making adaptive and validated modeling essential for this task. 
To address these, we propose an analytical framework (Figure~\ref{fig:main}) that translates collective decision-making observations into a structured causal grounding system that is inspectable, evolvable, and traceable. It adopts Codified Decision Tree (CDT)~\citep{peng2026cdt} as its structured behavior model, leveraging its inherent interpretability and extensibility.
Additionally, we introduce two components for advanced analysis: (1)~an \textbf{adapter mechanism} that adds, updates, and removes gates and statements in the CDT, supporting both time-aware evolution and group-aware transfer; and (2)~\textbf{traceable evidence nodes} that ground each node in the derived historical events, providing provenance from behavioral rules to predicted actions.

Our experiments split the benchmark chronologically, using the earlier portion for modeling and the later portion for prediction. Results show that our method outperforms trivial prompting and data-driven baselines such as Retrieval-Augmented Generation (RAG) across all dimensions. We further analyze this task from two perspectives. For \textbf{temporal analysis}, we reveal temporal behavioral drift within individual groups, and demonstrate that incrementally adapting the behavioral model to capture this drift yields stronger prediction performance without the cost of full retraining, offering practical use at scale. For \textbf{cross-group analysis}, we discover that groups within the same domain share similar behavioral patterns. This motivates cross-group knowledge transfer via the group-aware adapter, enabling more effective behavioral simulation for data-scarce groups.


In summary, our contributions are as follows:
\begin{itemize}[leftmargin=*]
    \item We formalize the task of \textbf{Organized Group Behavior Simulation} and construct \textbf{GROVE}, a benchmark from 8,052 real-world decision records covering 44 entities across 9 domains, as well as a multi-dimensional evaluation criterion.
    \item We propose a structured framework that translates historical events into interpretable, evolvable behavior model with adaptation mechanisms and traceable evidence, and results show that our method outperforms other baselines across all evaluation dimensions on GROVE.
    \item Our analysis reveals temporal behavioral drift within individual groups, which the time-aware adapter effectively captures to yield stronger prediction without full retraining, and cross-group similarity that enables knowledge transfer for data-scarce organizations.
\end{itemize}
    \section{Related Work}

\paragraph{Group Behavior Analysis}
Social science research has long established that organized groups are distinct decision-making entities whose actions cannot be reduced to those of individual members \citep{cyert1963behavioral, march1958organizations}. Organizations are characterized as boundedly rational actors following satisficing routines that evolve incrementally along path-dependent trajectories shaped by internal coalitions and institutional constraints \citep{north1990institutions, nelson1982evolutionary}.
Computational simulation of group-level dynamics has proven indispensable for strategic forecasting, policy evaluation, and organizational design \citep{epstein1996growing, davis2007developing}. 
Other work uses hierarchical clustering over users' temporal activity to characterize behaviorally distinct social-media groups~\citep{sinha2020hierarchical}.
While these foundations are well established, the emergence of large language models opens new opportunities to operationalize group behavior with richer cognitive fidelity.

\paragraph{LLMs for Behavior Simulation}
The emergence of LLMs has transformed behavior simulation, initially by modeling individual agents. \citet{park2023generative} demonstrate that agents equipped with memory and planning produce emergent social behaviors, while \citet{horton2023large} shows that LLMs can effectively simulate economic agents. This paradigm has been extended to collective settings, including political deliberations \citep{bobrov2025fomc, baker2024senate} and large-scale societal interactions \citep{piao2025agentsociety}. However, in these systems, group-level outcomes emerge from individual interactions, with no explicit representation of the group’s own decision-making logic. 
In contrast, we model the group itself as the primary unit of analysis, equipping it with an explicit, structured decision-making logic derived from historical records\citep{peng2026cdt}.
Beyond prediction, we further investigate adaptation mechanisms, cross-group behavioral similarity, temporal evolution of decision patterns, and cross-group knowledge transfer.

    \section{Framework}

\subsection{Task Formalization}
\label{sec:formalization}

We formalize \textbf{Organized Group Behavior Simulation} as follows.
Each organizational decision is represented as a context--decision pair $(c_i, d_i)$, where $c_i$ describes the situation the organization faces and $d_i$ is its observed response.
An organization's decision history is a chronological sequence $\mathcal{D} = \{(c_i, d_i)\}_{i=1}^{N}$.
The task is to induce a behavioral model $\mathcal{M}$ from historical decisions $\mathcal{D}_{\leq t}$ that can simulate the organization's response to a new situation:
\begin{equation}
    \hat{d} = \mathcal{M}(c^* \mid \mathcal{D}_{\leq t}),
    \label{eq:task}
\end{equation}
where $c^*$ is a new context occurring after time $t$ and $\hat{d}$ is the simulated decision.
Beyond point prediction, we require $\mathcal{M}$ to be \emph{inspectable}, \emph{evolvable}, and \emph{traceable}.

\subsection{Preliminary Knowledge: Codified Decision Tree}
\label{sec:cdt}

We adopt the Codified Decision Tree (CDT)~\citep{peng2026cdt} as the backbone
of our behavioral model.
The full construction is shown in Algorithm~\ref{alg:cdt-construction} and implementation details can be found in Appendix~\ref{sec:construction}.

\textbf{Structure.} As shown in Figure~\ref{fig:main}, a CDT $\mathcal{T}$ is a rooted tree where internal nodes contain gates $g$, natural-language conditions describing situational categories (e.g., ``the event involves regulatory compliance''), and all nodes maintain behavioral statements $\{s\}$ summarizing the organization's typical responses under those conditions (e.g., ``tends to act aggressively to regulatory uncertainty'').

\textbf{Inference Traversal.} Given a context $c$, the tree is traversed top-down: at each internal node, an LLM evaluates whether the gate condition $g$ holds for $c$. If true, the traversal continues into the corresponding subtree; otherwise, the subtree is skipped. The routing is non-exclusive, meaning a context may satisfy multiple gates at the same level.
The gate conditions along all activated paths and the statements from all reached nodes are collected and concatenated as background knowledge. This background, together with the context $c$, is provided to the LLM as input, and the model is instructed to predict the organization's next decision $\hat{d}$.

\begin{algorithm}[t]
\caption{Recursive CDT Construction.}
\label{alg:cdt-construction}
\footnotesize

\begin{algorithmic}[1]

\Require Observations $\mathcal{D}_n$; depth $d$; ancestor statements
$\mathcal{A}_n$;
\Ensure CDT subtree rooted at node $n$

\Function{BuildCDT}{$\mathcal{D}_n,d,\mathcal{A}_n$}

  \State Initialize node $n$

  \If{$|\mathcal{D}_n|<\tau_{\min}$ or $d>D_{\max}$} \Comment{$\tau_{\min}$: minimum samples; $D_{\max}$: maximum depth }
    \State \Return $n$
  \EndIf

  \State $\mathcal{C}\gets
    \Call{MultiPerspectiveCluster}{\mathcal{D}_n,R,m}$ \Comment{$R$: perspectives; $m$: clusters per perspective}

  \State $\mathcal{H}\gets
    \Call{GenerateHypotheses}{\mathcal{C},\mathcal{A}_n}$

  \State $\mathcal{H}\gets
    \Call{Deduplicate}{\mathcal{H}}$

  \ForAll{$(\hat{g},\hat{s})\in\mathcal{H}$} \Comment{$(\hat{g},\hat{s})$: gate--statement pair}

    \State $(status,\mathcal{D}_g)\gets
      \Call{Validate}{\hat{g},\hat{s},\mathcal{D}_n}$

    \If{$status=\textsc{Accept}$}
      \State Attach $\hat{s}$ to $n$

    \ElsIf{$status=\textsc{Recurse}$}
      \State $n'\gets
        \Call{BuildCDT}
        {\mathcal{D}_g,d+1,\mathcal{A}_n\cup\{\hat{s}\}}$

      \State Attach gate--child pair $(\hat{g},n')$ to $n$
    \EndIf

  \EndFor

  \State \Return $n$

\EndFunction
\end{algorithmic}
\end{algorithm}
\subsection{Traceable Evidence}
Each node $n$ in the CDT maintains a grounding matrix $\mathbf{M}_n \in \{\textsc{Sup}, \textsc{Con}, \textsc{Irr}\}^{|\mathcal{R}(n)| \times |\mathcal{S}_n|}$, where $\mathcal{R}(n)$ is the set of events routed to $n$ and $\mathcal{S}_n$ is the set of statements at $n$. Each entry $m_{ij}$ records whether event $e_i$ supports, contradicts, or is irrelevant to statement $s_j$, as determined by an NLI-based consistency check. This matrix serves as the evidence backbone of the framework: it grounds every behavioral statement in specific historical events, and drives the adaptation process (\S\ref{sec:adaptation}) by providing the empirical basis for deciding which statements to keep, add, demote, or delete. As a result, every node in the CDT is traceable to its originating historical events, providing end-to-end provenance from source evidence to predicted actions.

\subsection{Behavior Adaptation}
\label{sec:adaptation}
When new data $\mathcal{D}_{\text{new}}$ arise from temporal updates, cross-group integration, or domain-specific refinements, an adaptation process is triggered. Rather than rebuilding the tree from scratch, the process incrementally updates the existing CDT to incorporate the new information while preserving previously validated patterns.

The adaptation process operates top-down, processing each node $n$ sequentially based on the NLI scores of $\mathcal{D} \cup \mathcal{D}_{\text{new}}$. For each node, the algorithm evaluates every existing statement $s_j$ using its precision $p_j = |\mathrm{sup}_j| / (|\mathrm{sup}_j| + |\mathrm{con}_j|)$ and effective sample size $n_j = |\mathrm{sup}_j| + |\mathrm{con}_j|$, then generates new statements to cover remaining events. The following four operations are applied in order:
\begin{itemize}[leftmargin=*]
    \item \textbf{Keep}: Statements with $p_j \geq \tau_{\text{keep}}$ are empirically validated and retained. Statements with $n_j < \tau_{\min}$ are also retained due to insufficient evidence for a reliable judgment.
    \item \textbf{Delete}: Statements with $p_j < \tau_{\text{delete}}$ are predominantly contradicted and removed.
    \item \textbf{Demote}: Statements with $\tau_{\text{delete}} \leq p_j < \tau_{\text{keep}}$ hold for some observations but not others. The algorithm relocates each such statement to an existing child whose gate already routes its supporting events; if no suitable child exists, a new child node with a new gate is created.
    \item \textbf{Add}: Events not supported by any surviving statement form the uncovered set $\mathcal{U}_n$. When $|\mathcal{U}_n| \geq \tau_{\min}$, an LLM generates candidate statements conditioned on the gate path from the root to $n$. Each candidate is validated against all events at the node via the grounding matrix and accepted only if it meets the precision threshold $\tau_{\text{keep}}$.
\end{itemize}
After all four operations at node $n$ are completed, the algorithm recurses into each child node, passing down any demoted statements. This incremental, top-down design ensures that the CDT evolves with minimal modifications while maintaining interpretability and evidence grounding.

\subsection{Behavior Similarity Analysis Tools}
\label{sec:analysis-tool}
We analyze behavioral similarity at two levels using sentence embeddings $\mathbf{e}(\cdot)$: an event level that compares actions under analogous contexts, and a structural level that compares the CDTs.

\textbf{Event level.} We define a behavioral similarity score (BSS) to quantify whether two groups behave similarly under analogous contexts. Given two sets of context--action pairs from different temporal phases or different groups, the score identifies the top-$N$ event pairs whose context similarity exceeds a threshold $\tau$ and computes the average action similarity over these matched pairs:
\begin{equation}
    \text{BSS}(G_i, G_j) = \frac{1}{N} \sum_{(k, l) \in \text{Top-}N} \mathrm{sim}(a^i_k, a^j_l), \quad \left(\cos(\mathbf{e}_{c^i_k}, \mathbf{e}_{c^j_l}) > \tau\right)
\end{equation}
This score serves as the unified metric for both within-group temporal and cross-group similarity analysis throughout this paper.

\textbf{Structural level.} We directly compare CDTs $T_i$ and $T_j$ using Earth Mover's Distance (EMD)~\citep{710701} over their node embeddings:
\begin{equation}
    \text{EMD}_{\eta}(T_i, T_j) = \min_{F \geq 0} \sum_{k,l} f_{kl} \cdot \left(1 - \cos\!\left(\mathbf{e}^{\eta}_{k}(T_i), \mathbf{e}^{\eta}_{l}(T_j)\right)\right),
\end{equation}
where $F$ is a transport plan with uniform marginals and $\eta \in \{\text{gate}, \text{stmt}\}$ selects between gate embeddings (decision conditions) and statement embeddings (behavioral tendencies).
\section{GROVE Benchmark}
\label{sec:task}

We introduce GROVE, a benchmark for evaluating group-level behavioral prediction. It comprises two data sources: (1) \textbf{Wikipedia}, covering 35 groups across 9 domains with long-term historical records, and (2) \textbf{TechCrunch}, covering 9 technology corporations with recent news (January 2025--March 2026). Detailed information and examples are provided in Appendix~\ref{app:exp_dataset}.

\begin{figure}[t] 
    \centering
    \includegraphics[width=\linewidth]{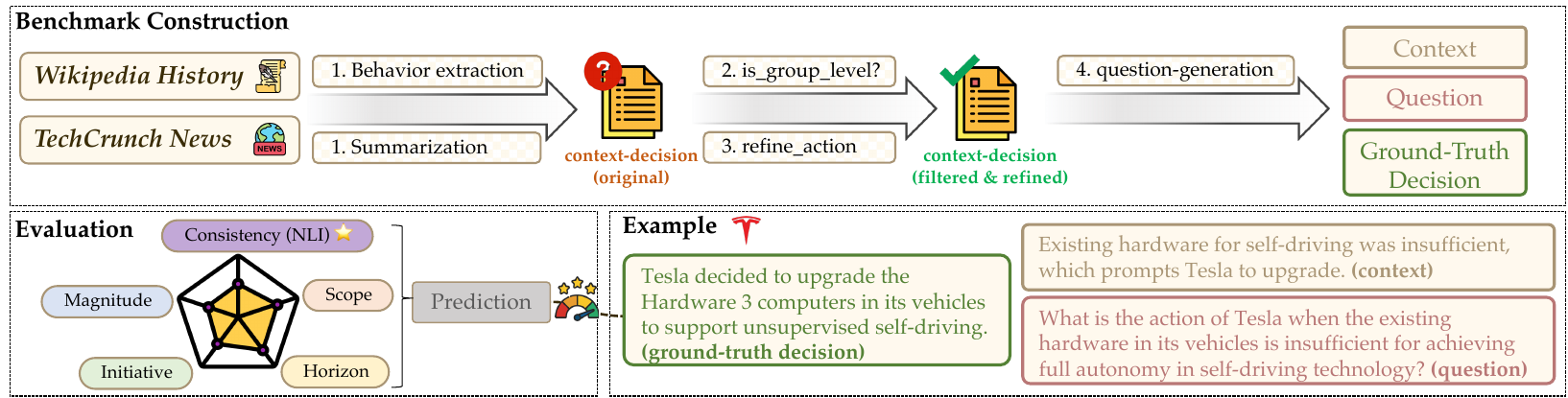} 
    \caption{Curation and Evaluation of GROVE Benchmark}
    \label{fig:benchmark}
\end{figure}

\subsection{Data Curation}

We prompt GPT-4.1 to extract group actions from Wikipedia narrative segments and to summarize background and behavior from TechCrunch news snippets (see Figure~\ref{fig:benchmark}). Since group behavior in historical records is often carried out by specific individuals, we design three filtering criteria grounded in organizational theory~\citep{cyert1963behavioral, march1958organizations} along the dimensions of attribution, succession, and significance (detailed in Appendix~\ref{app:cons_tests}), and prompt GPT-4.1 to filter out individual-level actions and refine the remaining ones to reflect group-level decision-making. After filtering, we retain only groups with at least 100 context--decision pairs to ensure sufficient data for behavioral model construction and evaluation. Each context--decision pair is further augmented with a guiding question that specifies the decision context (e.g., strategic domain or time frame) without revealing the actual~outcome. Full pipeline can be found in Appendix~\ref{sec:grove-construction}.

\subsection{Evaluation Criterion}
\label{sec:prediction}

During evaluation, the simulation model receives context and a guiding question, and predicts the group's next decision. Each prediction is compared against the reference behavior using an LLM-based evaluation suite across five dimensions~\citep{miles1978organizational, march1991exploration}: (1)~\textbf{consistency}, measuring entailment via NLI, scored as 100 (entailed), 50 (neutral), or 0 (contradicts); (2)~\textbf{initiative}, measuring strategic posture (proactive, neutral, or reactive); (3)~\textbf{scope}, measuring orientation (internal or external); (4)~\textbf{magnitude}, measuring response intensity (incremental, moderate, or transformative); and (5)~\textbf{horizon}, measuring temporal orientation (exploitative, balanced, or explorative). For dimensions (2)--(5), matched classifications score 100 and mismatches score 0. Human validation of benchmark curation quality and evaluation reliability demonstrates strong agreement between human and LLM judge, with full results provided in Appendix~\ref{sec:human_validation}.

    \section{Experiments and Analysis}
\subsection{Experimental Setup}
We evaluate our method against five baselines. \textbf{Vanilla} prompts the LLM with only the context and question, without any grounding information. \textbf{LoRA Finetuning}~\citep{hu2022lora} adapts the base model by parameter-efficient finetuning on historical data. \textbf{Human Profile} provides a human-written group description adapted from Wikipedia as grounding. \textbf{Summarization-based}~\citep{yuan2024evaluating} extracts and summarizes historical records into a unified textual behavior profile appended to the prompt. \textbf{RAG}~\citep{10.5555/3495724.3496517} retrieves the most similar past context--decision examples from training data and appends them to the prompt.
For all experiments, we split the data chronologically into 70\% train and 30\% test for both Wikipedia and TechCrunch. We use Qwen2.5-7B-Instruct as the prediction model and GPT-4.1 as the evaluation model in the main experiments. Detailed hyperparameter setups and model selection in the framework are placed in Appendix~\ref{app:exp-details}. Additional experiments across seven other models and a configuration threshold sensitivity study are reported in Appendix~\ref{sec:expanded-model-results} and \ref{sec:threshold-sensitivity}.





\begin{figure*}[t] 
    \centering
    \begin{minipage}[c]{0.35\textwidth} 
        \centering
        \includegraphics[width=0.9\linewidth]{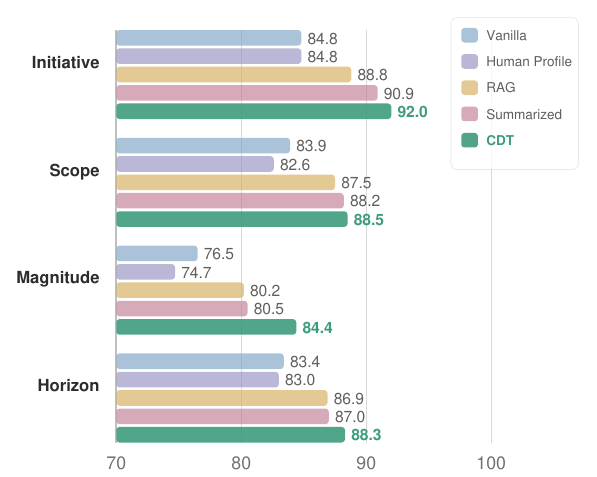}
        \vspace{-5pt} 
        \caption{Multi-dimensional evaluation results.}
        \label{fig:5-dim}
    \end{minipage}
    \hfill
    \begin{minipage}[c]{0.63\textwidth}
        \centering
        \definecolor{bestgreen}{RGB}{198,239,206}
        
        \resizebox{\linewidth}{!}{
            \begin{tabular}{@{}l cccccccccc | c@{}}
                \toprule
                & \multicolumn{10}{c|}{\textbf{Wikipedia}} & \textbf{TC} \\
                \cmidrule(lr){2-11}
                \textbf{Method} & Aero. & Auto. & Broad. & Edu. & Ener. & Game & Ret. & Spo. & Tech. & \textit{Avg} & \textit{Avg} \\
                \midrule
                Vanilla          & 67.7 & 62.9 & 64.4 & 69.4 & 52.1 & 59.2 & 61.1 & 54.4 & 62.4 & 61.5 & 78.5 \\
                LoRA Finetuning & 54.8 & 52.8 & 52.2 & 56.2 & 57.3 & 50.0 & 48.6 & 43.4 & 59.2 & 52.7 & 67.6 \\
                Human Profile    & \cellcolor{bestgreen}\textbf{70.7} & 65.4 & 63.1 & 68.1 & 55.2 & 62.5 & 55.6 & 53.9 & 63.9 & 62.0 & 79.8 \\
                Summarized-based & 69.8 & 62.9 & 66.2 & 68.1 & 68.8 & \cellcolor{bestgreen}\textbf{65.0} & 59.7 & 58.3 & 64.9 & 64.9 & 79.0 \\
                RAG              & 67.3 & 62.5 & 64.4 & 66.0 & 60.4 & 59.2 & 55.6 & 55.2 & 67.5 & 62.0 & 84.2 \\
                Ours (CDT)       & 70.0 & \cellcolor{bestgreen}\textbf{69.1} & \cellcolor{bestgreen}\textbf{68.5} & \cellcolor{bestgreen}\textbf{70.8} & \cellcolor{bestgreen}\textbf{74.0} & \cellcolor{bestgreen}\textbf{65.0} & \cellcolor{bestgreen}\textbf{66.7} & \cellcolor{bestgreen}\textbf{64.5} & \cellcolor{bestgreen}\textbf{70.1} & \cellcolor{bestgreen}\textbf{68.7} & \cellcolor{bestgreen}\textbf{85.7} \\
                \midrule
                \multicolumn{12}{l}{Temporal Adaptation Study evaluation on P2 \& P3} \\
                \midrule
                Fixed CDT        & 70.2 & 65.0 & 69.0 & 67.7 & 75.0 & 62.5 & \cellcolor{bestgreen}\textbf{66.7} & 65.2 & \cellcolor{bestgreen}\textbf{75.8} & 68.6 & -- \\
                Retrained CDT    & 71.2 & 67.5 & 73.4 & 66.7 & 73.4 & 62.5 & 54.2 & 66.7 & 67.2 & 67.0 & -- \\
                Adapted CDT      & \cellcolor{bestgreen}\textbf{76.0} & \cellcolor{bestgreen}\textbf{68.3} & \cellcolor{bestgreen}\textbf{77.2} & \cellcolor{bestgreen}\textbf{75.0} & \cellcolor{bestgreen}\textbf{78.1} & \cellcolor{bestgreen}\textbf{72.5} & 64.6 & \cellcolor{bestgreen}\textbf{68.8} & 68.8 & \cellcolor{bestgreen}\textbf{72.1} & -- \\
                \bottomrule
            \end{tabular}%
        }
        \captionof{table}{Consistency results on GROVE benchmark.}
        \label{tab:main_results}
    \end{minipage}
\end{figure*}

\subsection{Behavior Prediction Performance}
\label{sec:main-results}
\begin{tcolorbox}[
    colback=yellow!10,
    colframe=yellow!50!black,
    fontupper=\normalsize,
    arc=3pt,
    boxrule=0.8pt,
    left=3pt, right=3pt, top=3pt, bottom=3pt
]
\textbf{\textit{Takeaway.}} 
Group behavior simulation benefits more from the structured and controllable representation provided by our CDT than from relying solely on textual summaries or retrieved historical records.
\end{tcolorbox}

As shown in Table~\ref{tab:main_results} and Figure~\ref{fig:5-dim}, our framework achieves the best overall scores across all five evaluation dimensions, surpassing Vanilla, Human Profile, Summarization-based, and RAG baselines. 
RAG tends to over-rely on the most similar past events without sufficient generalization, making its predictions particularly vulnerable to temporal behavioral drift and leading to significant performance degradation in later evaluation phases. Summarization-based provides noticeable gains over Vanilla and Human Profile in several settings, confirming the value of grounding predictions in past behavioral histories, but it remains clearly behind our method, which benefits from a structured, controllable behavioral representation rather than pure textual aggregation. LoRA Finetuning underperforms Vanilla, which can be attributed to repeating similar decisions in the train split rather than learning global behavior pattern.
Results using GPT-4o-mini as the prediction model are in Appendix~\ref{app:results}.

\subsection{Temporal Behavioral Drift}
\begin{tcolorbox}[
    colback=yellow!10,
    colframe=yellow!50!black,
    fontupper=\normalsize,
    arc=3pt,
    boxrule=0.8pt,
    left=3pt, right=3pt, top=3pt, bottom=3pt
]
\textbf{\textit{Takeaway.}} 
Organized group behavior can be analyzed at the event level across temporal phases, revealing how decision-making patterns evolve over time. 
Our dynamic mechanism incrementally tracks these strategic shifts by continuously updating the behavioral model.
\end{tcolorbox}

\label{sec:temporal_drift}
\begin{figure}[h]
    \centering
    \includegraphics[width=\linewidth]{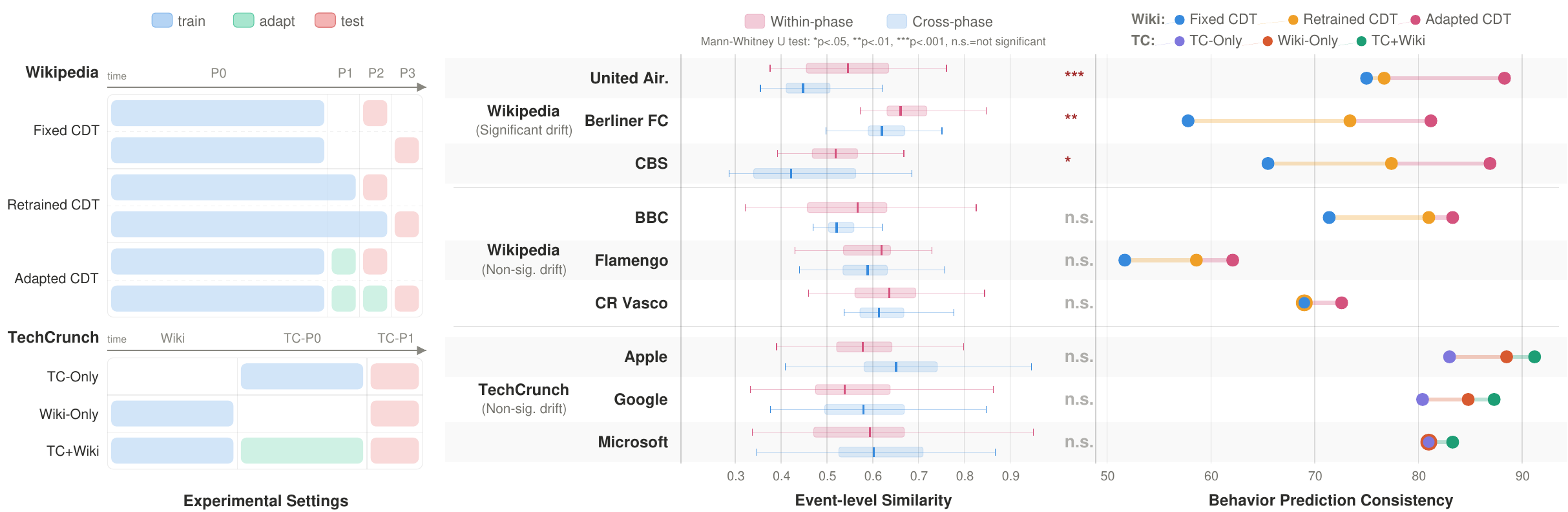}
    \captionof{figure}{Adaptation settings and results for groups with significant drift ($*$), non-significant drift (n.s.), and cross-source transfer from Wikipedia to TechCrunch.}
    \label{fig:drift_performance}
\end{figure}

\paragraph{Pre-Analysis.} Behavior patterns of organized groups are expected to shift over decades or a century decision-making history. We first examine this at the event level. By splitting test set into three phases chronologically, we calculate the behavior similarity score (BSS) of within- and cross-phase. Results show that over 57\% of groups in the Wikipedia exhibit significantly higher within-phase similarity than cross-phase similarity (p $<$ 0.05, Mann-Whitney U test), providing model-agnostic evidence of widespread behavioral drift directly in the raw data. As shown in Figure~\ref{fig:drift_performance}, groups with significant drift (e.g., United Airlines, Berliner FC, CBS) exhibit clear separation between within-phase and cross-phase similarity, while groups with non-significant drift (e.g., BBC, Flamengo, CR Vasco) show overlapping distributions. Moreover, all 9 TechCrunch groups show no significant drift, as their records span only one year, confirming that drift is a long-horizon phenomenon. Full analysis for each group can be found in Appendix~\ref{app:results}.

\paragraph{Experiments.} Since behavioral drift exists, a static CDT will progressively lose fidelity, motivating the need for adaptation. To address this, we leverage time-aware adapter on previous base CDT. A case can be found in Section~\ref{sec:case}. As shown in Figure~\ref{fig:drift_performance}, we use the three-phase data split and compare several settings: Fixed CDT, Retrained CDT, and Adapted CDT.
We also evaluate cross-source transfer by adapting Wikipedia-built CDTs on TechCrunch cutoff data. As shown in Table~\ref{tab:main_results}, the Adapted CDT outperforms the Fixed CDT on consistency in most domains, and even surpasses Retrained CDT. This is because the adaptation mechanism inherently increases the weight of the most recent data in the model while preserving previously validated behavioral rules, resulting in more accurate modeling of near-future behavior. Notably, as shown in Figure~\ref{fig:drift_performance}, groups that exhibit more significant behavioral drift in the event-level analysis also show larger performance gains from adaptation, further supporting that the adapter's gains are driven by capturing temporal drift. Moreover, for cross-source transfer, TC+Wiki consistently outperforms TC-Only, demonstrating that the adapter effectively bridges the distributional gap between different data sources. These results demonstrate that our method effectively updates the behavioral model in response to temporal drift, achieving stronger predictive performance without the cost of full retraining.

\subsection{Group Similarity and Knowledge Transfer}
\begin{tcolorbox}[
    colback=yellow!10,
    colframe=yellow!50!black,
    fontupper=\normalsize,
    arc=3pt,
    boxrule=0.8pt,
    left=3pt, right=3pt, top=3pt, bottom=3pt
]
\textbf{\textit{Takeaway.}} 
Groups within the same domain tend to converge on structurally similar behavioral patterns, observable at both the event and structural levels. This shared behavioral pattern can be exploited through group-aware transfer adaptation, providing an effective approach for modeling data-scarce groups.
\end{tcolorbox}

\begin{figure*}[h]
\centering
\begin{minipage}[l]{0.45\textwidth}
  \centering
  \includegraphics[width=\linewidth]{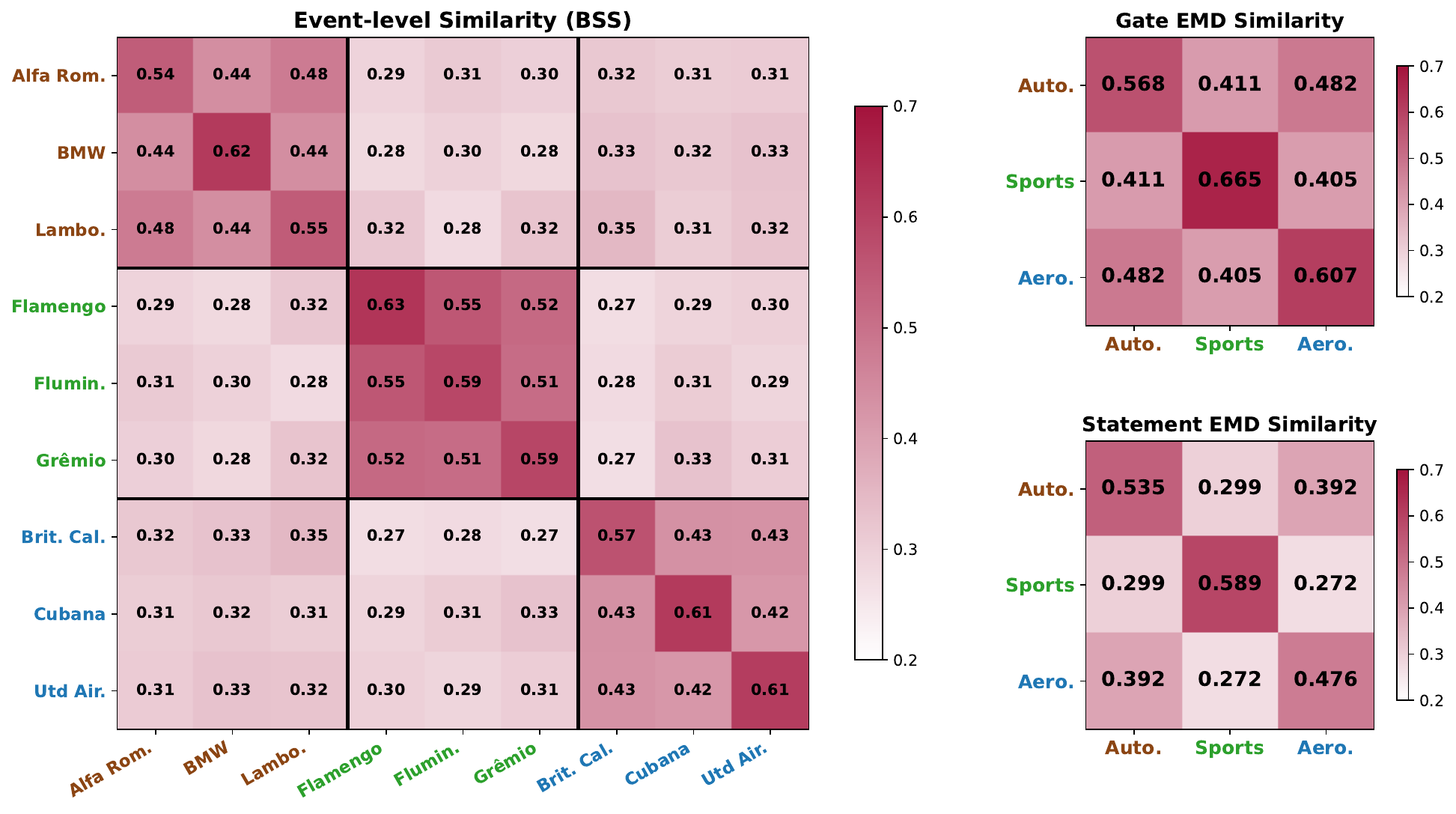}
  \captionof{figure}{Group similarity analysis.}
  \label{fig:group_similarity}
\end{minipage}
\hfill
\begin{minipage}[r]{0.54\textwidth}
  \centering
  \resizebox{\linewidth}{!}{
  \begin{tabular}{lllccc}
    \toprule
    Domain & Source & Target & Vanilla & Tgt CDT & Transfer \\
    \midrule
    Aero.
      & Brit. Cal. & Airbus & 75.0 & 61.4 & \textbf{83.3} \\
    \midrule           
    Auto.
      & Volkswagen & Ford & 73.8 & 66.7  &  \textbf{76.2} \\    
    \midrule
    \multirow{2}{*}{Spo.}
      & Flumin. & Derry City & 65.0 & 71.7 & \textbf{80.0} \\
      & SE\_Palmeiras & GELP        & 66.7 & 64.6 & \textbf{68.8} \\
    \midrule
    \multirow{4}{*}{Edu.}
      & Cornell & Princeton  & 65.4 & 76.9 & \textbf{84.6} \\
      & Cornell & UPenn      & 53.3 & 60.0 & \textbf{63.3} \\
      & Cornell & Stanford   & 79.4 & 70.6 & \textbf{82.4} \\
      & GaTech  & UCLA       & 62.5 & 62.5 & \textbf{70.0} \\
    \bottomrule
  \end{tabular}
  }
  \captionof{table}{CDT Transfer Experiment on Multiple Sources Compared to Vanilla and Target CDT.}
  \label{tab:transfer}
\end{minipage}
\end{figure*}

\textbf{Group Similarity} can be expected since groups operating within the same domain naturally face similar market forces, regulatory environments, and competitive dynamics. We first investigate this at two levels. As shown in Figure~\ref{fig:group_similarity} for automobile, sports and aerospace, intra-domain similarity consistently exceeds inter-domain similarity at the event-level. At the structural level, EMD-based similarity (Section~\ref{sec:analysis-tool}) matrix shows that groups within the same domain have higher similarity than cross-domain in both gates and statements. Beyond these selected groups, these patterns hold across all domains and detailed results can be found in Figure~\ref{fig:similarity-full-1}.

\textbf{Knowledge Transfer.} Both event-level and structural-level analyses reveal behavioral similarity among groups within the same domain, which raises a practical question: can behavioral knowledge be transferred from data-rich groups to improve simulation for data-scarce ones? As shown in Table~\ref{tab:transfer}, directly applying a CDT built from a data-scarce group's limited data performs poorly, sometimes even falling below the vanilla setting, because insufficient data cannot support the construction of a well-structured behavioral model. 
By leveraging shared behavioral structures inherent within a domain, our adaptation mechanism enables the CDT to maintain robust predictive performance even under data sparsity.
This ensures high-fidelity simulation in practical settings where historical records are unavailable or limited, such as for newly formed organizations, thereby expanding the applicability of behavioral modeling to a wider range of real-world entities.

    \section{Case Studies}
\label{sec:case}
\begin{figure*}[h]
    \centering
    \includegraphics[width=\linewidth]{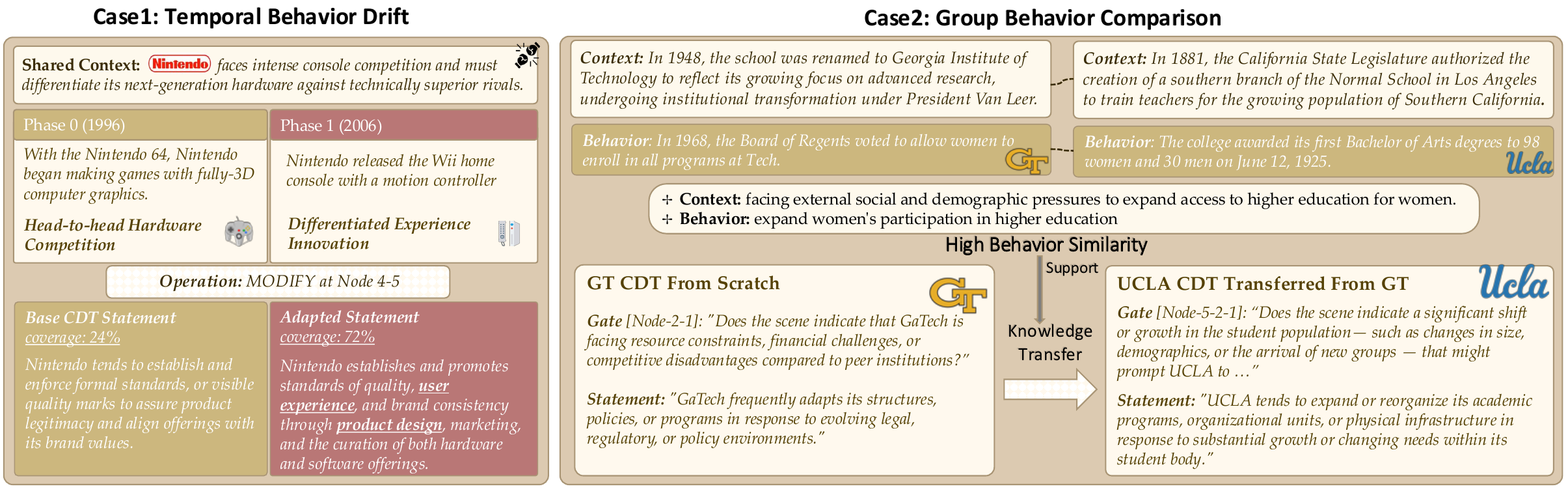}
    \caption{Cases for temporal behavior drift (left), group similarity and knowledge transfer (right) with traceable evidence node for validation}
    \label{fig:case_studies}
\end{figure*}

As shown in Figure~\ref{fig:case_studies}, the left case shows that temporal behavioral drift happens when facing similar competitive pressure but making fundamentally different decisions, which supports the necessity to modify the CDT to fit behavior better. The right case shows that universities within the same domain share similar behavioral patterns regarding women's admission (e.g., GaTech and UCLA independently responded through similar institutional mechanisms), which supports transferring GaTech's behavioral model to data-scarce UCLA. These highlight the advantages of our framework: the ability to capture temporal behavioral drift through the time-aware adapter, and to identify cross-group similarity and enable knowledge transfer through the group-aware adapter. Moreover, traceable evidence nodes in our framework ensure that both the modeling and adaptation processes are anchored in verifiable historical events, making every analytical result explainable and auditable. CDT structure and traversal cases are provided in Appendix~\ref{app:cases}.
    \section{Conclusion}
We formalize Organized Group Behavior Simulation and construct GROVE, a benchmark with 8,052 real-world context--decision pairs across 9 domains and a multi-dimensional evaluation protocol. We propose a framework that converts historical observations into an inspectable, evolvable behavioral model with adaptation mechanisms and traceable evidence nodes, outperforming summarization- and retrieval-based baselines across all evaluation dimensions. Our analysis reveals temporal behavioral drift within individual groups, effectively captured by the time-aware adapter without full retraining, and structured cross-group similarity that enables knowledge transfer for data-scarce organizations.
As future work, we plan to extend toward multi-agent organizational simulation where interacting groups co-evolve their strategies.
    \section*{Acknowledgments}
This work is sponsored in part by NSF CAREER Award 2239440, NSF Proto-OKN Award 2333790, Sponsored Research Projects from companies like Cisco and eBay, as well as generous gifts from Google, Adobe, and Teradata. Any opinions, findings, and conclusions or recommendations expressed herein are those of the authors and should not be interpreted as necessarily representing the views, either expressed or implied, of the U.S. Government. The U.S. Government is authorized to reproduce and distribute reprints for government purposes not withstanding any copyright annotation hereon.
    \bibliography{colm2026_conference}
    \bibliographystyle{colm2026_conference}

    \newpage
    \appendix
    \section{Framework Details}

\subsection{Codified Decision Tree Construction}
\label{sec:construction}

Given an organization's decision history $\mathcal{D} = \{(c_i, d_i)\}_{i=1}^{N}$, we construct a \emph{Codified Decision Tree} (CDT) that organizes behavioral patterns into a hierarchy of interpretable, conditional rules.
A CDT is a rooted tree in which each node $n$ stores a set of behavioral statements $\mathcal{S}_n$ that generalize the organization's tendencies in natural language, together with a set of gate--child pairs $\mathcal{G}_n = \{(g_k, n_k)\}_{k=1}^{K_n}$. Each gate $g_k$ is a yes/no question about the observable context and $n_k$ is the corresponding child node.
At inference time the tree is traversed top-down: root statements are collected unconditionally; for each gate satisfied by the input context (judged by a discriminator LLM), the traversal recurses into the child, accumulating progressively more specific statements that serve as background knowledge for the prediction LLM.

All prompt templates used in construction and adaptation are provided in Appendix~\ref{app:prompts}. Construction proceeds recursively. A node becomes a leaf when the number of routed observations falls below a minimum or the depth exceeds $D_{\max}$. Otherwise, the following four-step pipeline is executed. All hyperparameters are listed in Table~\ref{tab:hyperparams}.

The complete construction can be found in Algorithm~\ref{alg:cdt-construction}; this section provides the corresponding implementation details.

\paragraph{Step 1: Multi-perspective clustering.}
We perform $R$ rounds of KMeans clustering, each re-embedding the entire observation set under a different behavioral lens.
Four guide suffixes are appended to each context $c_i$ before encoding, priming the embedding toward distinct dimensions:
\emph{``Thus, $\mathcal{O}$ will''} (decision verbs),
\emph{``Thus, $\mathcal{O}$ prioritizes''} (goal nouns),
\emph{``As a result, $\mathcal{O}$ faces''} (pressures and risks),
and \emph{``This changes $\mathcal{O}$'s''} (strategic domains),
where $\mathcal{O}$ denotes the organization name.
In each round $r$, we compute a composite embedding per observation:
\begin{equation}
\mathbf{e}_i^{(r)} = \left[\frac{\mathrm{GenEnc}(c_i \oplus \mathrm{suffix}_r)}{\|\mathrm{GenEnc}(c_i \oplus \mathrm{suffix}_r)\|_2} ;\ \frac{\mathrm{SurfEnc}(d_i)}{\|\mathrm{SurfEnc}(d_i)\|_2}\right]
\end{equation}
where $\mathrm{GenEnc}$ extracts the last-token hidden state of Qwen3-8B applied to the suffix-augmented context and $\mathrm{SurfEnc}$ encodes the decision via OpenAI's \texttt{text-embedding-3-small}; both are $\ell_2$-normalized before concatenation.
KMeans is then applied, and for each centroid the $m$ nearest observations form a representative cluster. Accumulating clusters across all rounds ensures coverage of complementary behavioral dimensions.

\paragraph{Step 2: Hypothesis generation.}
For each cluster $\mathcal{C}_j$, a generator LLM proposes $k$ candidate behavioral statements (decision hypotheses) and $k$ corresponding context-trigger questions (gate hypotheses) via a chain-of-thought prompt.
The prompt supplies the cluster's context--decision pairs, all statements already established at ancestor and current nodes, and the gate path from the root.
The LLM is instructed to identify the main behavioral feature not yet captured, formulate $k$ concise grounding statements, and propose $k$ selective yes/no context-trigger questions focused on the organization's next decision.

\paragraph{Step 3: Summarization and deduplication.}
Raw hypotheses from all clusters are aggregated and may overlap. An LLM-based compression step merges redundant pairs, rewrites them for generality, and retains a small set of meaningfully distinct statement--gate pairs.

\paragraph{Step 4: Two-stage validation.}
Each surviving pair $(\hat{g}, \hat{s})$ is validated against the full observation set $\mathcal{D}_n$ at the current node using a discriminator LLM.

\emph{Stage 1 --- Ungated validation.}
The statement $\hat{s}$ is checked against every observation without any gate filter (see Appendix~\ref{app:prompts} for prompt templates).
The ungated precision is $p_{\mathrm{global}} = N_{\mathrm{yes}} / (N_{\mathrm{yes}} + N_{\mathrm{no}})$. If $p_{\mathrm{global}} \geq \tau_{\mathrm{accept}}$, the statement is added unconditionally to $\mathcal{S}_n$ and the gate is discarded.

\emph{Stage 2 --- Gated validation.}
Remaining hypotheses undergo gate filtering. Each observation is evaluated against the gate $\hat{g}$ by the discriminator, and only those answered ``yes'' are retained as $\mathcal{D}_n^{\hat{g}}$. The gated precision $p_{\mathrm{gated}}$ is computed on this subset, and the broadness $b = 1 - N_{\mathrm{irr}} / |\mathcal{D}_n|$ measures gate selectivity. A gate is accepted only if $b \leq \tau_{\mathrm{filter}}$. If $p_{\mathrm{gated}} \geq \tau_{\mathrm{accept}}$, the statement is installed as a leaf child with no further recursion. If $\tau_{\mathrm{reject}} \leq p_{\mathrm{gated}} < \tau_{\mathrm{accept}}$, the filtered observations are passed to a new child for recursive construction. Otherwise, the hypothesis is discarded.

\subsection{Demotion Details}
\label{sec:demotion-details}

As described in \S\ref{sec:adaptation}, a statement $s_j$ is demoted when $\tau_{\text{delete}} \leq p_j < \tau_{\text{keep}}$, indicating that it holds for some observations but not others.
The demotion procedure attempts to identify a subcontext where the statement achieves adequate precision and relocate it accordingly.
All statements marked for demotion at the same node are processed jointly in three steps.

\paragraph{Step~1: Route through existing children.}
For each demoted statement $s_n^{(j)}$, we route its supporting events $\mathrm{sup}_j$ through the existing child gates of $n$.
If a single child $n_c$ captures $\geq 50\%$ of $\mathrm{sup}_j$, we move $s_n^{(j)}$ to $n_c$, provided the statement's precision recomputed at $n_c$ is $\geq \tau_{\text{keep}}$.

\paragraph{Step~2: Hypothesize new gates.}
For statements not resolved in Step~1, we group their supporting and contradicting events and prompt an LLM to hypothesize a gate condition that separates $\mathrm{sup}_j$ from $\mathrm{con}_j$.
The candidate gate is accepted only if $\geq 50\%$ of $\mathrm{sup}_j$ are routed to the new child and the statement's precision under that child is $\geq \tau_{\text{keep}}$.
If both criteria are met, a new child node is created under $n$ with the hypothesized gate, and the statement is placed there.
When multiple statements at the same node require new gates, the LLM may propose a shared gate if their supporting events overlap substantially.

\paragraph{Step~3: Fallback.}
Statements for which neither Step~1 nor Step~2 succeeds are deleted, as no subcontext with adequate precision can be found.

    \newpage
\section{Benchmark and Groups Details}

\label{app:exp_dataset}

\subsection{End-to-End GROVE Construction Pipeline}
\label{sec:grove-construction}

GROVE is constructed through the following end-to-end pipeline.
\begin{enumerate}[leftmargin=*]
  \item \textbf{Data collection.} We collect section-level historical
    narratives for 35 organized groups across 9 domains from Wikipedia. For
    recent corporate behavior, we collect TechCrunch articles concerning 9
    technology corporations from January 2025 through March 2026.
  \item \textbf{Behavior extraction.} For Wikipedia, GPT-4.1 identifies
    candidate actions taken by the focal group from each narrative segment.
    For TechCrunch, GPT-4.1 converts each article into a structured
    $(\text{background},\text{action})$ pair, where the background describes
    the situation preceding the focal corporate action.
  \item \textbf{Group-behavior filtering.} Each candidate is evaluated using
    the attribution, successor, and significance tests in
    Appendix~\ref{app:cons_tests}. The filter removes private individual
    actions, actions that do not represent the organization's authority or
    institutional continuity, and actions without organization-level
    significance.
  \item \textbf{Action refinement.} For each retained candidate, the action is
    rewritten with the organized group as the subject while preserving the
    original event semantics. This normalization prevents incidental
    individual names from obscuring the collective actor represented by the
    benchmark.
  \item \textbf{Question generation and leakage control.} GPT-4.1 generates a
    guiding question that specifies the relevant decision dimension without
    revealing the reference action. We then apply empty-field filtering,
    action deduplication, explicit action-span removal from the background,
    and lexical-overlap containment checks to reduce answer leakage.
  \item \textbf{Retention and chronological splitting.} We retain groups with
    at least 100 surviving context--decision pairs. The final benchmark
    contains 44 entities and 8,052 pairs. Each group's records are ordered
    chronologically and split into 70\% training and 30\% test data, yielding
    5,386 training and 2,666 test pairs in total.
\end{enumerate}
For reproducibility, the release includes the curation and evaluation prompts,
model versions, random seeds, filtering criteria, threshold settings, and code
used to produce the benchmark.

\begin{table*}[h]
    \centering
    \small
    \begin{tabular}{llrr|llrr}
    \toprule
    \textbf{Dom.} & \textbf{Group} & \textbf{Tr} & \textbf{Te} & \textbf{Dom.} & \textbf{Group} & \textbf{Tr} & \textbf{Te} \\
    \midrule
    Aero. & Boeing & 111 & 48 & Spo.* & CR Vasco & 148 & 64 \\
     & Brit. Caledonian & 195 & 84 &  & Dallas Cowboys & 142 & 61 \\
     & Cubana & 99 & 43 &  & Bayern Munich & 72 & 32 \\
     & United Airlines & 103 & 45 &  & Fluminense & 296 & 127 \\
    Auto. & Alfa Romeo & 133 & 58 &  & Grêmio & 129 & 56 \\
     & Chrysler & 105 & 46 &  & McLaren & 180 & 78 \\
     & GM & 117 & 51 &  & Anderlecht & 75 & 33 \\
     & Lamborghini & 84 & 37 &  & Real Madrid & 76 & 33 \\
     & Volkswagen & 81 & 36 &  & Red Bull Racing & 93 & 41 \\
    Broad. & ABC & 121 & 52 &  & Palmeiras & 191 & 83 \\
     & BBC & 72 & 32 &  & UFC & 100 & 43 \\
     & CBS & 147 & 64 & Tech. & Apple & 126 & 55 \\
     & CNN & 70 & 31 &  & Microsoft & 96 & 42 \\
     & Australian BC & 100 & 43 & TC & Amazon & 102 & 72 \\
    Edu. & Cornell & 75 & 33 &  & Anthropic & 94 & 64 \\
     & Georgia Tech & 89 & 39 &  & Apple & 180 & 121 \\
    Ener. & ExxonMobil & 109 & 48 &  & Google & 274 & 184 \\
    Game & Nintendo & 140 & 60 &  & Meta & 192 & 132 \\
    Ret. & Target & 84 & 36 &  & Microsoft & 82 & 56 \\
    Spo. & Al Ahly SC & 100 & 44 &  & Nvidia & 64 & 45 \\
     & Berliner FC & 114 & 49 &  & OpenAI & 252 & 170 \\
     & CR Flamengo & 101 & 44 &  & Tesla & 72 & 51 \\
    \midrule
    \multicolumn{8}{l}{\textbf{Total: 44 groups, 5386 train, 2666 test}} \\
    \bottomrule
    \end{tabular}
    \caption{Experiment dataset statistics.}
    \label{tab:exp_stats}
\end{table*}

\subsection{Benchmark Construction Tests}
\label{app:cons_tests}

\paragraph{Attribution Test.}
The attribution test verifies that the candidate action can be reasonably attributed to the organization as a collective actor, rather than being solely a private or incidental act of an individual. For example, a CEO announcing a company-wide restructuring plan is attributable to the corporation, whereas the same CEO's personal real estate purchase is not. This test filters out actions that merely involve members of the organization but do not represent the organization's collective agency, ensuring that the benchmark reflects decisions made by the group rather than decisions made near the group.

\paragraph{Successor Test.}
The successor test addresses cases where an action was initiated or carried out by a specific individual but nonetheless represents the organization's interests, authority, or institutional capacity. Even if a single leader made the decision, it qualifies as group behavior if it was later absorbed, continued, or institutionalized by the organization. For instance, a university president unilaterally establishing a new research center counts as organizational behavior if the institution subsequently maintained and expanded the center beyond that president's tenure. This test captures the organizational theory insight that institutions routinely act through designated agents whose decisions bind and persist beyond the individual~\citep{march1958organizations}.

\paragraph{Significance Test.}
The significance test requires that the candidate action has a meaningful impact at the level of the organization, not merely on the individual actor. Routine personal decisions by group members, even when made in an organizational context, are excluded if they do not materially affect the group's trajectory, reputation, or operations. For example, an employee's promotion within a company is typically insignificant at the organizational level, whereas a major product launch or a strategic acquisition is. This test ensures that the benchmark focuses on decisions that shape the organization's behavioral trajectory, which is the level at which our simulation task operates.

\begin{figure}[h]
\begin{tcolorbox}[colback=gray!5, colframe=gray!50, fontupper=\small\ttfamily, left=4pt, right=4pt, top=4pt, bottom=4pt]
\# History\\
\{history\}\\[4pt]
\# Candidate Action\\
\{action\}\\[4pt]
\# Task\\
Determine whether the candidate action should be considered a valid \textbf{group behavior} of \textbf{\{group\_name\}}.\\[2pt]
Answer \textbf{yes} only if ALL of the following conditions are satisfied:\\[2pt]
1. \textbf{Attribution Test}\\
\hspace*{1em}The action can be reasonably attributed to \{group\_name\} as a collective actor, rather than being solely a private or incidental act of an individual.\\[2pt]
2. \textbf{Successor Test}\\
\hspace*{1em}Even if the action was initiated by an individual, it plausibly represents the interests, authority, or capacity of \{group\_name\}, or was later institutionalized by \{group\_name\}.\\[2pt]
3. \textbf{Significance Test}\\
\hspace*{1em}The action has a meaningful impact at the level of \{group\_name\}, not merely on the individual actor.\\[2pt]
If \textbf{any} condition is not satisfied, answer \textbf{no}.\\[4pt]
\{``reason'': ``<brief explanation>'',\\
\ ``is\_group\_behavior'': ``yes'' | ``no''\}
\end{tcolorbox}
\caption{Group behavior filtering prompt with three criteria: attribution, successor, and significance tests.}
\label{fig:prompt-filter}
\end{figure}
\FloatBarrier

    \newpage
\section{Human Validation}
\label{sec:human_validation}

To verify the reliability of the LLM-assisted benchmark curation and the
consistency judge, we conduct a two-part human validation study covering
(1)~benchmark data quality and (2)~judge--human agreement.

\subsection{Benchmark Quality Validation}
\label{sec:bench_quality}

\paragraph{Setup.}
We stratified-sample 100 entries from the Wikipedia-sourced benchmark across
all 9 domains and 50 entries from the TechCrunch-sourced benchmark. Two
primary annotators independently label every item. For each Wikipedia entry,
they answer three binary (Yes/No) questions:
\begin{itemize}[nosep]
  \item \textbf{Q1 (Group Behavior)}: Is the extracted action a valid group
    behavior attributable to the organization as a collective actor?
  \item \textbf{Q2 (Rewrite Fidelity)}: Is the rewritten action faithful to
    the original---i.e., only the subject is changed, with no semantic
    distortion?
  \item \textbf{Q3 (Question Quality)}: Is the generated evaluation question
    reasonable and does it avoid leaking the ground-truth action?
\end{itemize}
For each TechCrunch entry, they answer:
\begin{itemize}[nosep]
  \item \textbf{Q4 (Summary Accuracy)}: Is the GPT-generated background an
    accurate and faithful summary of the article context?
  \item \textbf{Q5 (No Leakage)}: Does the background avoid revealing or
    implying the extracted action?
\end{itemize}

Agreement and Cohen's $\kappa$ are computed between the two primary
annotators. Whenever they disagree, a third annotator reviews that item and
selects the final label. Thus, an agreed item uses the shared primary label,
whereas a disagreed item uses the third annotator's label; this is equivalent
to a three-person majority decision while requiring the third annotator only
for disagreements. Agreement between the final human label and GPT-4.1 is
computed over \emph{all} sampled items, not only the subset on which the two
primary annotators agree.

\paragraph{Results.}
Table~\ref{tab:bench_quality} reports the human agreement rates and
inter-annotator reliability. Across all five questions, the two primary
annotators show substantial to near-perfect agreement
($\kappa=0.76$--$0.91$), and the final human labels agree with the GPT-4.1
curation decisions on 90--97\% of the samples. Rewrite fidelity achieves the
highest agreement ($\kappa=0.91$), indicating that subject rewriting rarely
changes the action semantics. Group-behavior validity has slightly lower but
still substantial agreement ($\kappa=0.76$), with disagreements concentrated
in borderline cases where an individual's action was later institutionalized
by the organization. The TechCrunch results likewise support the faithfulness
and leakage control of the generated backgrounds.

\begin{table}[H]
\centering
\small
\begin{tabular}{llccc}
\toprule
\textbf{Source} & \textbf{Question} & \textbf{Agree (\%)} &
\textbf{$\kappa$} & \textbf{Final Human--GPT (\%)} \\
\midrule
\multirow{3}{*}{Wikipedia}
  & Q1: Group Behavior    & 89.0 & 0.76 & 92.0 \\
  & Q2: Rewrite Fidelity  & 96.0 & 0.91 & 97.0 \\
  & Q3: Question Quality  & 93.0 & 0.82 & 94.0 \\
\midrule
\multirow{2}{*}{TechCrunch}
  & Q4: Summary Accuracy  & 92.0 & 0.78 & 90.0 \\
  & Q5: No Leakage        & 94.0 & 0.84 & 96.0 \\
\bottomrule
\end{tabular}
\caption{Benchmark quality validation. ``Agree'' is agreement between the two
primary annotators. ``Final Human--GPT'' is agreement between the adjudicated
human label and GPT-4.1 over all sampled items.}
\label{tab:bench_quality}
\end{table}

\subsection{Consistency-Judge Agreement Validation}
\label{sec:judge_agreement}

\paragraph{Setup.}
To validate the GPT-4.1 consistency judge, we sample another 100
prediction--reference pairs, approximately balanced across Entail, Neutral,
and Contradict and drawn from both Wikipedia and TechCrunch results under
multiple inference settings. Two primary annotators independently assign one
of the three labels using the same rubric as the GPT judge:
\begin{itemize}[nosep]
  \item \textbf{Entail}: The reference entails the prediction and follows the
    same group logic.
  \item \textbf{Neutral}: The reference is neutral to the prediction and
    reflects a different facet.
  \item \textbf{Contradict}: The reference contradicts the prediction and
    follows an opposing group logic.
\end{itemize}
As in the benchmark-quality study, a third annotator resolves every primary
disagreement. Human--human statistics use the two primary annotations, while
human--GPT statistics compare GPT-4.1 against the final adjudicated label over
all 100 pairs.

\paragraph{Results.}
Table~\ref{tab:judge_agreement} summarizes the agreement statistics. The two
primary annotators achieve 78.0\% exact agreement with Cohen's $\kappa=0.67$
and Spearman $\rho=0.74$. Comparing the final human label with GPT-4.1 gives
73.0\% exact agreement, $\kappa=0.60$, and Spearman $\rho=0.71$. The GPT
judge's agreement with the final human label is therefore close to the
human--human agreement under this constrained consistency rubric. The
per-class breakdown in Table~\ref{tab:judge_agreement} shows the highest
agreement for Entail and the lowest for Neutral, reflecting the greater
subjectivity of deciding whether two actions describe different facets.

\begin{table}[H]
\centering
\small
\begin{tabular}{lccc}
\toprule
\textbf{Comparison} & \textbf{Agree (\%)} & \textbf{$\kappa$} &
\textbf{Spearman $\rho$} \\
\midrule
H--H (primary annotators) & 78.0 & 0.67 & 0.74 \\
Final H--GPT-4.1         & 73.0 & 0.60 & 0.71 \\
\bottomrule
\end{tabular}
\par\smallskip
\begin{tabular}{l|ccc|c}
\toprule
\textbf{Final Human} & \textbf{Entail} & \textbf{Neutral} &
\textbf{Contradict} & \textbf{Agree (\%)} \\
\midrule
\textbf{Entail}     & 27 & 5  & 1  & 81.8 \\
\textbf{Neutral}    & 4  & 21 & 9  & 61.8 \\
\textbf{Contradict} & 2  & 6  & 25 & 75.8 \\
\bottomrule
\end{tabular}
\caption{Consistency-judge validation. H--H compares the two primary
annotators; Final H--GPT compares the adjudicated human label against GPT-4.1
over all 100 pairs. The lower block gives per-class Final H--GPT agreement.}
\label{tab:judge_agreement}
\end{table}

    \newpage
\section{Case Studies}
\label{app:cases}

\begin{figure}[H]
    \centering
    \includegraphics[width=\linewidth]{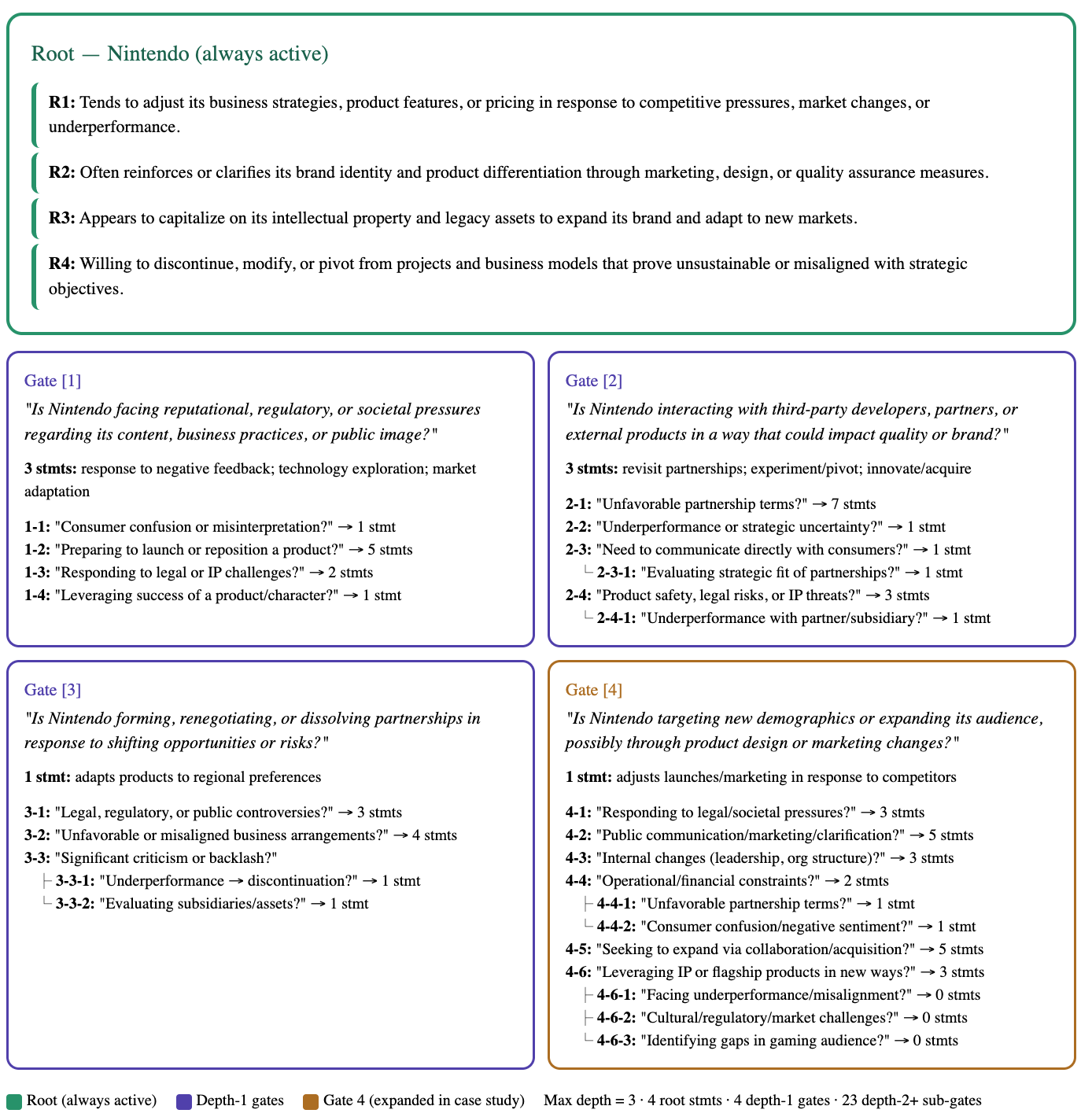}
    \caption{CDT Visualization for Nintendo}
    \label{fig:cdt-case}
\end{figure}

\begin{figure}[H]
    \centering
    \includegraphics[width=0.9\linewidth]{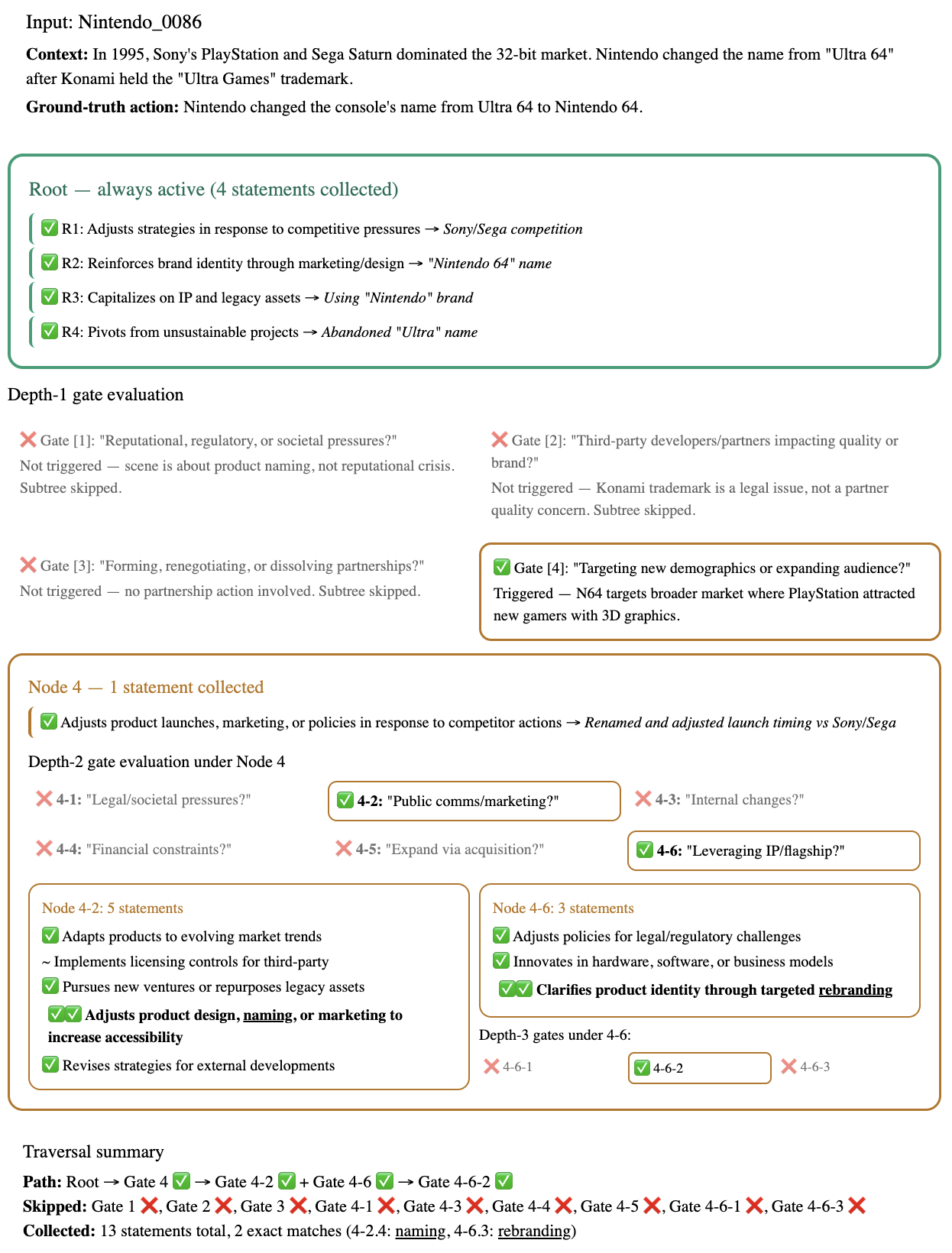}
    \caption{A running case for one data sample from Nintendo}
    \label{fig:running-case}
\end{figure}
    \newpage
\section{Setup Details}
\label{app:exp-details}

\subsection{Hyperparameters}
\label{sec:hyperparams}
Table~\ref{tab:hyperparams} summarizes all hyperparameters used in CDT construction and adaptation.
\begin{table}[h]
\centering
\small
\begin{tabular}{llc}
\toprule
\textbf{Symbol} & \textbf{Description} & \textbf{Value} \\
\midrule
$D_{\max}$            & Maximum tree depth                        & 3    \\
$R$                   & Clustering rounds (guide suffixes)         & 4    \\
$m$                   & Observations sampled per centroid          & 8    \\
$k$                   & Hypotheses generated per cluster           & 3    \\
\midrule
$\tau_{\mathrm{accept}}$ / $\tau_{\mathrm{keep}}$  & Precision threshold to accept / keep a statement  & 0.65 \\
$\tau_{\mathrm{reject}}$ / $\tau_{\text{delete}}$ & Precision threshold to reject / delete a statement & 0.35 \\
$\tau_{\mathrm{filter}}$ & Broadness upper bound for gate acceptance  & 0.8  \\
$\tau_{\min}$            & Minimum sample size for evaluation         & 3    \\
\bottomrule
\end{tabular}
\caption{Hyperparameters for CDT construction (\S\ref{sec:construction}) and adaptation (\S\ref{sec:adaptation}). Thresholds $\tau_{\mathrm{accept}}$ and $\tau_{\mathrm{keep}}$ share the same value, as do $\tau_{\mathrm{reject}}$ and $\tau_{\text{delete}}$.}
\label{tab:hyperparams}
\end{table}

\subsection{Baselines.}
\textbf{Human profiles} - craft a short paragraph which introduces basic knowledge of this group without leaking potential actions and their results. Figure ~\ref{fig:human-profile} shows human profile of Boeing.

\begin{figure}[h]                                                                                                                                                      
  \begin{tcolorbox}[colback=gray!5, colframe=gray!50, fontupper=\small\ttfamily, left=4pt, right=4pt, top=4pt, bottom=4pt]                                               
  \# Human Profile\\ 
  The Boeing Company is an American multinational corporation that designs, manufactures, and sells airplanes, rotorcraft, rockets, satellites, and missiles worldwide.  
  The company also provides leasing and product support services. Boeing is among the largest global aerospace manufacturers; it is the fourth-largest defense contractor
   in the world based on 2022 revenue and is the largest exporter in the United States by dollar value. Boeing was founded in 1916 by William E. Boeing in Seattle, Washington.\\[4pt]
  As of 2023, the Boeing Company's corporate headquarters is located in the Crystal City neighborhood of Arlington County, Virginia. The company is organized into three
  primary divisions: Boeing Commercial Airplanes (BCA), Boeing Defense, Space \& Security (BDS), and Boeing Global Services (BGS). In 2021, Boeing recorded \$62.3       
  billion in sales. Boeing is ranked 54th on the Fortune 500 list (2020), and ranked 121st on the Fortune Global 500 list (2020).
  \end{tcolorbox}        
  \caption{Example of a human-written Wikipedia profile used as background knowledge (Boeing).}
  \label{fig:human-profile}
  \end{figure}

\textbf{RAG}~\citep{10.5555/3495724.3496517} retrieves the 8 most similar context--decision pairs from the training data using OpenAI's text-embedding-large model, and appends them as in-context examples to the prompt. At inference time, the model is asked to predict the group's next action based on these retrieved examples and the current context. This approach directly leverages historical precedents but lacks any abstraction over behavioral patterns, making it sensitive to surface-level context similarity and vulnerable to temporal behavioral drift when the most similar past events no longer reflect the group's current decision-making logic.

\textbf{Summarization-based}~\citep{yuan2024evaluating} constructs a textual behavior profile for each group by prompting an LLM to generate a narrative-style summary from blocks of historical context--decision pairs. The prompt instructs the model to produce a cohesive profile weaving together the group's background, core behavioral patterns, key decisions, and strategic development, written in an informative style similar to a comprehensive organizational guide. The resulting profile is appended to the prediction prompt as grounding information. While this approach captures higher-level behavioral tendencies than RAG, the profile is a static, unstructured text that cannot be incrementally updated or decomposed for targeted analysis.

\textbf{LoRA Fine-tune} trains a separate Qwen2.5-7B-Instruct adapter for
each group using that group's initial historical training split (P0), which is
the same group-specific data available to the other behavior-modeling
baselines. We use LoRA rank $r=16$, scaling factor $\alpha=32$, dropout 0.05,
batch size 16, and AdamW with learning rate $2\times10^{-4}$.

\textbf{Model Selection.} Our framework employs different models for different roles based on their complexity requirements. For CDT construction and adaptation, GPT-4.1\footnote{\url{https://openai.com/index/gpt-4-1/}} serves as the main engine for proposing gates and statements, while GPT-4.1-mini handles lightweight classification tasks including gate traversal, coverage checking, and NLI-based relation matrix computation. All sentence embeddings for context matching and similarity analysis are computed using OpenAI's text-embedding-3-large\footnote{\url{https://openai.com/index/new-embedding-models-and-api-updates/}}. We use GPT-4.1 as the judge model for multi-dimensional evaluation scoring. For prediction, we use Qwen2.5-7B-Instruct\citep{Yang2024Qwen25TR} and GPT-4o-mini\citep{openai2024gpt4o}.

    \newpage
\section{Additional Experiments Results}
\label{app:results}

\subsection{Expanded Prediction Model Coverage}
\label{sec:expanded-model-results}

GROVE is designed primarily to compare behavior-modeling methods under a
controlled prediction backbone, rather than to rank language models by their
parametric knowledge. To test whether the observed benefit depends on the two
prediction models used in the original experiments, we repeat the comparison
with seven recent open- and closed-source models. Recent models may encode more
public organizational history in their parameters, creating a stronger
Vanilla baseline and a possible ceiling effect. Nevertheless, as shown in
Table~\ref{tab:expanded_model_coverage}, CDT obtains the best Wikipedia and
TechCrunch consistency scores for every evaluated prediction model.

\begin{table*}[h!]
\centering
\small
\resizebox{\textwidth}{!}{%
\begin{tabular}{lccccc}
\toprule
\textbf{Prediction Model} & \textbf{Vanilla} & \textbf{Human Profile} &
\textbf{Summarization} & \textbf{RAG} & \textbf{Ours (CDT)} \\
\midrule
Qwen3-4B-Instruct-2507 & 57.9 / 79.8 & 59.4 / 83.0 & 64.6 / 86.2 &
  56.8 / 83.8 & \textbf{68.2 / 86.6} \\
Qwen3.5-4B & 55.0 / 82.9 & 53.8 / 81.7 & 58.7 / 85.7 &
  58.1 / 81.3 & \textbf{63.1 / 88.3} \\
gpt-5-nano-2025-08-07 & 76.1 / 88.8 & 78.1 / 88.8 & 77.7 / 90.5 &
  76.2 / 88.5 & \textbf{83.1 / 91.6} \\
gpt-5-mini-2025-08-07 & 80.4 / 89.8 & 80.2 / 90.1 & 81.6 / 92.1 &
  77.9 / 86.0 & \textbf{84.4 / 93.0} \\
gpt-5-2025-08-07 & 83.0 / 89.7 & 82.8 / 89.9 & 84.2 / 90.3 &
  81.5 / 85.2 & \textbf{87.2 / 92.0} \\
gemini-3-flash-preview & 64.0 / 85.9 & 67.4 / 86.2 & 78.0 / 88.0 &
  74.1 / 75.9 & \textbf{79.2 / 89.4} \\
gemini-3.1-flash-lite & 65.3 / 89.2 & 62.9 / 87.4 & 73.1 / 89.1 &
  73.0 / 80.1 & \textbf{77.7 / 91.8} \\
\bottomrule
\end{tabular}%
}
\caption{Consistency results across prediction models. Each entry reports
Wikipedia / TechCrunch average.}
\label{tab:expanded_model_coverage}
\end{table*}

The gains persist for the strongest prediction models: relative to Vanilla,
CDT improves gpt-5-mini by 4.0 / 3.2 points and gpt-5 by 4.2 / 2.3 points on
Wikipedia / TechCrunch. The same qualitative result holds for the Qwen and
Gemini families, indicating that the improvement is not restricted to an
OpenAI prediction backbone or to small models.

\subsubsection{Temporal Adaptation Across Prediction Models}
We further repeat the temporal adaptation experiment with the same seven
prediction models. Table~\ref{tab:expanded_temporal_adaptation} shows that the
Adapted CDT obtains the highest average consistency score for every model,
supporting the benefit of time-aware structural updates across model families.

\begin{table}[h!]
\centering
\small
\resizebox{\columnwidth}{!}{%
\begin{tabular}{lccc}
\toprule
\textbf{Prediction Model} & \textbf{Fixed CDT} & \textbf{Retrained CDT} &
\textbf{Adapted CDT} \\
\midrule
Qwen3-4B-Instruct-2507 & 66.2 & 66.6 & \textbf{68.5} \\
Qwen3.5-4B            & 61.1 & 61.4 & \textbf{65.1} \\
gpt-5-nano-2025-08-07 & 80.1 & 82.0 & \textbf{86.0} \\
gpt-5-mini-2025-08-07 & 83.6 & 83.4 & \textbf{85.3} \\
gpt-5-2025-08-07      & 85.5 & 86.1 & \textbf{88.3} \\
gemini-3-flash-preview & 79.8 & 79.7 & \textbf{80.1} \\
gemini-3.1-flash-lite  & 77.8 & 76.1 & \textbf{80.3} \\
\bottomrule
\end{tabular}%
}
\caption{Temporal adaptation consistency results across prediction models.}
\label{tab:expanded_temporal_adaptation}
\end{table}

\subsection{Threshold Sensitivity}
\label{sec:threshold-sensitivity}

The construction thresholds $\tau_{\mathrm{accept}}$ and
$\tau_{\mathrm{reject}}$ and the adaptation thresholds
$\tau_{\mathrm{keep}}$ and $\tau_{\mathrm{delete}}$ denote the same precision
cutoffs at different pipeline stages. A statement is accepted or kept when it
has sufficient support, and rejected or deleted when it is predominantly
contradicted; consequently, we share their values. Pilot experiments indicated
stable behavior over a reasonable range, so we use $\tau_{\mathrm{accept}}=\tau_{\mathrm{keep}}=0.65$ and
$\tau_{\mathrm{reject}}=\tau_{\mathrm{delete}}=0.35$, close to the setting in
prior CDT work~\citep{peng2026cdt}, as a representative configuration rather
than a tuned optimum.

For sensitivity analysis, we rebuild CDTs under nine threshold configurations
and evaluate five representative groups with three prediction models while
holding all other hyperparameters fixed. Table~\ref{tab:threshold_sensitivity}
shows no monotonic advantage from increasing or decreasing either cutoff. The
cross-model average ranges from 74.43 to 77.73; the default configuration
scores 76.33, within 1.40 points of the best configuration.

\begin{table*}[h!]
\centering
\small
\begin{tabular}{lccrrrr}
\toprule
\textbf{Config.} & $\boldsymbol{\tau_{\mathrm{accept}}}$ &
$\boldsymbol{\tau_{\mathrm{reject}}}$ & \textbf{gpt-5-nano} &
\textbf{gpt-5-mini} & \textbf{Qwen3-4B-Inst.} & \textbf{Average} \\
\midrule
C0 & 0.65 & 0.35 & 80.2 & 84.2 & 64.6 & 76.33 \\
C1 & 0.55 & 0.35 & 82.3 & \textbf{85.5} & 63.5 & 77.10 \\
C2 & 0.75 & 0.35 & 80.4 & 84.6 & 63.1 & 76.03 \\
C3 & 0.85 & 0.35 & 83.6 & 83.2 & 60.5 & 75.77 \\
C4 & 0.65 & 0.25 & \textbf{85.1} & 82.4 & \textbf{65.7} &
  \textbf{77.73} \\
C5 & 0.65 & 0.45 & 81.2 & 80.9 & 63.9 & 75.33 \\
C6 & 0.75 & 0.50 & 79.7 & 81.6 & 62.0 & 74.43 \\
C7 & 0.50 & 0.50 & 80.2 & 82.2 & 63.6 & 75.33 \\
C8 & 0.85 & 0.50 & 82.8 & 82.1 & 60.8 & 75.23 \\
\bottomrule
\end{tabular}
\caption{Threshold sensitivity on five representative groups. C0 is the
default configuration.}
\label{tab:threshold_sensitivity}
\end{table*}
\FloatBarrier

\begin{table*}[h!]   
      \centering
      \definecolor{bestgreen}{RGB}{198,239,206}                                                                                                                          
      \resizebox{\textwidth}{!}{%
      \begin{tabular}{@{}l ccccccccc | c | c@{}}
      \toprule                                                                                                                                                           
      & \multicolumn{9}{c|}{\textbf{Wikipedia}} & & \textbf{TC} \\
      \cmidrule(lr){2-10}                                                                                                                                                
      \textbf{Method}
      & Aero.
      & Auto.                                                                                                                                                            
      & Broad.    
      & Edu.                                                                                                                                                             
      & Ener.     
      & Game                                                                                                                                                             
      & Ret.      
      & Spo.
      & Tech.
      & \textit{Avg}
      & \textit{Avg} \\
      \midrule                                                                                                                                                           
      Vanilla                   & 80.0 & 80.7 & 79.8 & 78.7 & 86.5 & 74.2 & \cellcolor{bestgreen}\textbf{72.2} & 77.2 & \cellcolor{bestgreen}\textbf{82.7} & 79.1 & 89.9 
  \\                                                                                                                                                                     
      Human Profile             & 81.0 & 78.8 & 78.7 & 75.3 & 80.2 & 75.8 & 66.7 & 77.9 & 81.5 & 77.3 & 88.0 \\                                                          
      Summary-based             & 80.2 & 78.9 & 79.6 & 79.0 & 87.5 & 74.2 & 69.4 & 77.9 & 80.5 & 78.6 & 88.7 \\                                                          
      RAG                       & 77.8 & 74.5 & 80.9 & 80.1 & 85.4 & 77.5 & 69.4 & 75.0 & 75.6 & 77.4 & 85.9 \\                                                          
      Ours (CDT)                & \cellcolor{bestgreen}\textbf{82.8} & \cellcolor{bestgreen}\textbf{82.8} & \cellcolor{bestgreen}\textbf{85.8} &                         
  \cellcolor{bestgreen}\textbf{81.0} & \cellcolor{bestgreen}\textbf{92.7} & \cellcolor{bestgreen}\textbf{79.2} & 68.1 & \cellcolor{bestgreen}\textbf{82.7} & 78.9 &      
  \cellcolor{bestgreen}\textbf{81.6} & \cellcolor{bestgreen}\textbf{90.2} \\                                                                                             
      \midrule                                                                                                                                                           
      \textit{Temporal Adaptation Study} & & & & & & & & & & & \\                                                                                                        
      \midrule                                                                                                                                                           
      Ours (CDT)                & 82.1 & 81.5 & 85.3 & 77.4 & 82.8 & 85.0 & \cellcolor{bestgreen}\textbf{70.8} & 80.9 & 86.4 & 81.4 & -- \\                              
      Ours (CDT retrained)      & 82.7 & 82.1 & 85.7 & 76.3 & \cellcolor{bestgreen}\textbf{85.9} & 76.2 & 66.7 & \cellcolor{bestgreen}\textbf{82.3} &                    
  \cellcolor{bestgreen}\textbf{88.2} & 80.7 & -- \\                                                                                                                      
      Ours (CDT + Adaptation)   & \cellcolor{bestgreen}\textbf{83.8} & \cellcolor{bestgreen}\textbf{83.0} & \cellcolor{bestgreen}\textbf{87.3} &                         
  \cellcolor{bestgreen}\textbf{78.6} & \cellcolor{bestgreen}\textbf{85.9} & \cellcolor{bestgreen}\textbf{86.2} & 68.8 & 81.4 & 88.0 & \cellcolor{bestgreen}\textbf{82.6} 
  & -- \\         
      \bottomrule                                                                                                                                                        
      \end{tabular}                                                                                                                                                      
      }
      \caption{Consistency results on GROVE benchmark (gpt-4o-mini).}
      \label{tab:gpt4omini_results}
  \end{table*}

\begin{table*}[h!]                                                                                                                                                                              
      \centering
      \definecolor{bestgreen}{RGB}{198,239,206}                                                                                                                                                  
      \begin{tabular}{lllccc}                                                                                                                                                                    
      \toprule                                                                                                                                                                                   
      Domain & Source & Target & Vanilla & Tgt CDT & Transfer \\
      \midrule                                                                                                                                                                                   
      Aero.       
        & Brit. Cal. & Airbus & \cellcolor{bestgreen}\textbf{84.1} & 79.5 & 79.5 \\                                                                                                              
      \midrule                                                                                                                                                                                   
      Auto.                                                                                                                                                                                      
        & Volkswagen & Ford & 81.0 & 81.0 & \cellcolor{bestgreen}\textbf{88.1} \\                                                                                                                
      \midrule                                                                                                                                                                                   
      \multirow{2}{*}{Spo.}
        & Flumin. & Derry City & 80.0 & 80.0 & \cellcolor{bestgreen}\textbf{83.3} \\                                                                                                             
        & SE\_Palmeiras & GELP & \cellcolor{bestgreen}\textbf{87.5} & 70.8 & 85.4 \\                                                                                                             
      \midrule                                                                                                                                                                                   
      \multirow{4}{*}{Edu.}                                                                                                                                                                      
        & Cornell & Princeton & \cellcolor{bestgreen}\textbf{100.0} & 96.2 & 96.2 \\                                                                                                             
        & Cornell & UPenn & 66.7 & 76.7 & \cellcolor{bestgreen}\textbf{80.0} \\                                                                                                                  
        & Cornell & Stanford & 85.3 & 85.3 & \cellcolor{bestgreen}\textbf{88.2} \\                                                                                                               
        & GaTech & UCLA & 77.5 & \cellcolor{bestgreen}\textbf{80.0} & \cellcolor{bestgreen}\textbf{80.0} \\                                                                                      
      \bottomrule                                                                                                                                                                                
      \end{tabular}                                                                                                                                                                              
      \caption{CDT Transfer Experiment (gpt-4o-mini, Consistency Score)}
      \label{tab:transfer_gpt4omini_consistency}
  \end{table*}

\begin{table*}[h!]                                                                                                                                                                              
      \centering  
      \definecolor{bestgreen}{RGB}{198,239,206}                                                                                                                                                  
      \begin{tabular}{lllccc}                                                                                                                                                                    
      \toprule
      Domain & Source & Target & Vanilla & Tgt CDT & Transfer \\                                                                                                                                 
      \midrule                                                                                                                                                                                   
      Aero.
        & Brit. Cal. & Airbus & 71.6 & 62.5 & \cellcolor{bestgreen}\textbf{76.1} \\                                                                                                              
      \midrule                                                                                                                                                                                   
      Auto.
        & Volkswagen & Ford & 79.2 & 75.0 & \cellcolor{bestgreen}\textbf{82.1} \\                                                                                                                
      \midrule                                                                                                                                                                                   
      \multirow{2}{*}{Spo.}
        & Flumin. & Derry City & \cellcolor{bestgreen}\textbf{74.2} & 68.3 & 71.2 \\                                                                                                             
        & SE\_Palmeiras & GELP & \cellcolor{bestgreen}\textbf{79.2} & 60.9 & 72.9 \\                                                                                                             
      \midrule                                                                                                                                                                                   
      \multirow{4}{*}{Edu.}                                                                                                                                                                      
        & Cornell & Princeton & 74.0 & 76.9 & \cellcolor{bestgreen}\textbf{80.8} \\                                                                                                              
        & Cornell & UPenn & 81.7 & 77.5 & \cellcolor{bestgreen}\textbf{83.3} \\                                                                                                                  
        & Cornell & Stanford & \cellcolor{bestgreen}\textbf{75.0} & 66.9 & 72.8 \\                                                                                                               
        & GaTech & UCLA & 79.4 & 84.4 & \cellcolor{bestgreen}\textbf{85.6} \\                                                                                                                    
      \bottomrule                                                                                                                                                                                
      \end{tabular}                                                                                                                                                                              
      \caption{CDT Transfer Experiment (Qwen2.5-7B-Instruct, 4-Dim Score)}
      \label{tab:transfer_qwen_4dim}
  \end{table*} 

\begin{figure}[h!] 
    \centering
    \includegraphics[width=0.95\linewidth]{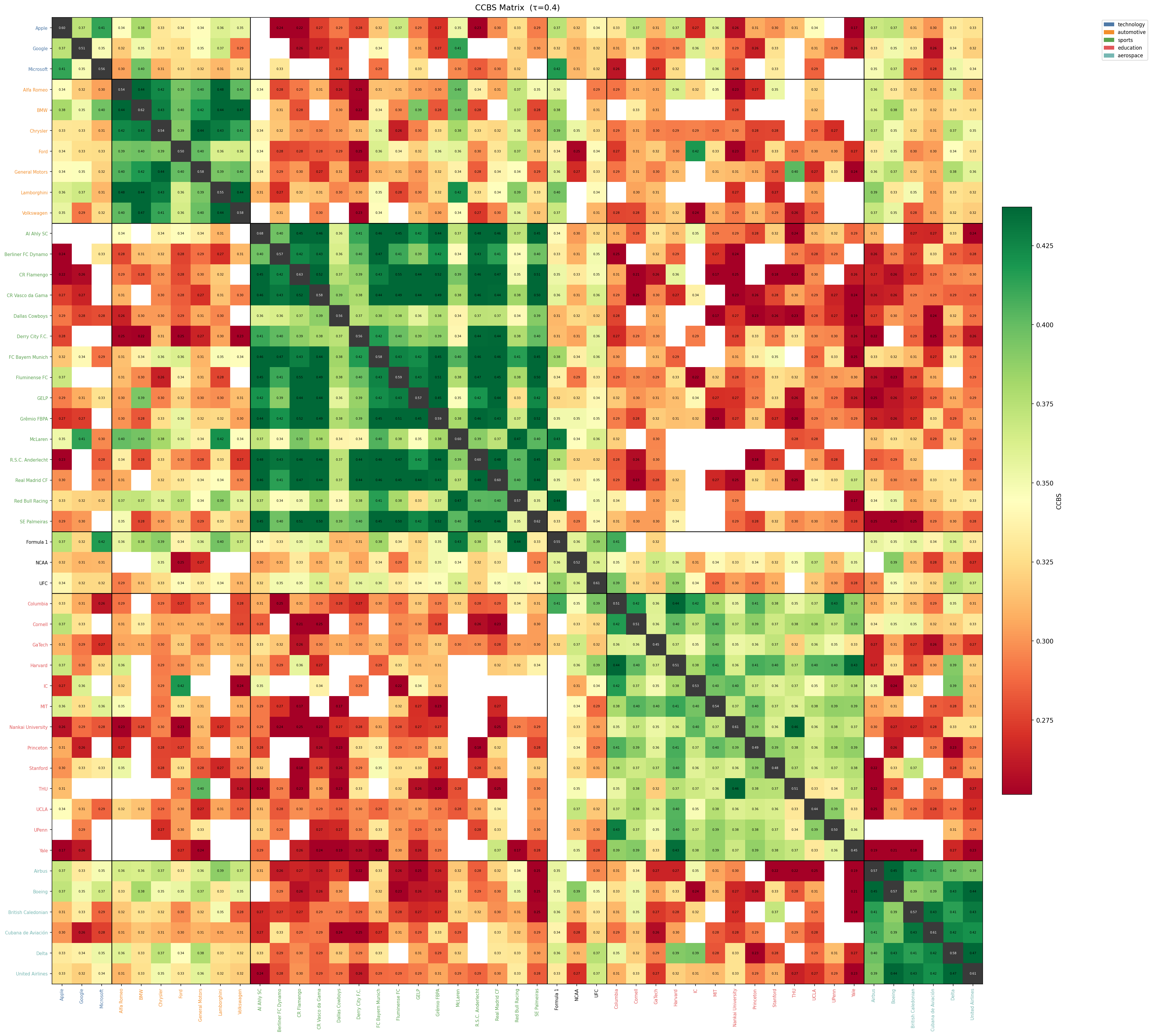} 
    \caption{Event-level Analysis for Group Similarity on Technology, Aerospace, Education, Sports and Automotive}
    \label{fig:similarity-full-1}
\end{figure}

\begin{figure}[h!] 
    \centering
    \includegraphics[width=0.95\linewidth]{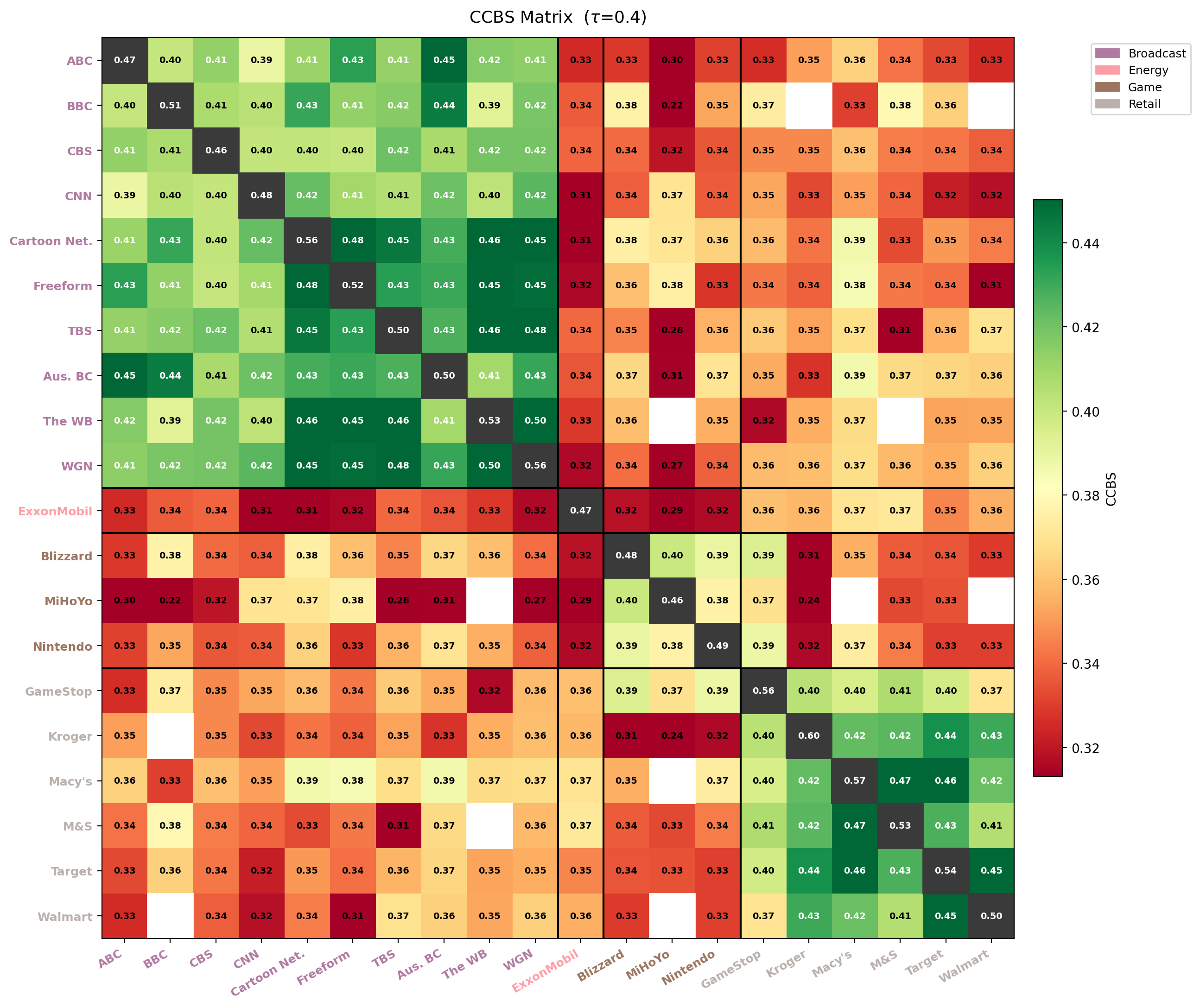} 
    \caption{Event-level Analysis for Group Similarity on Broadcast, Energy, Game, Retail}
    \label{fig:similarity-full-2}
\end{figure}

\begin{table*}[h!]
      \centering                        
      \definecolor{bestgreen}{RGB}{198,239,206}
      \resizebox{\textwidth}{!}{%
      \begin{tabular}{@{}l ccccccccc | c@{}}
      \toprule
      & \multicolumn{9}{c|}{\textbf{Wikipedia}} & \\
      \cmidrule(lr){2-10}
      \textbf{Method}
      & Aero. & Auto. & Broad. & Edu. & Ener. & Game & Ret. & Spo. & Tech.
      & \textit{Avg} \\
      \midrule    
      Ours (CDT)                & \cellcolor{bestgreen}\textbf{74.1} & \cellcolor{bestgreen}\textbf{77.1} & 74.4 & 74.8 & 76.6 & 75.6 & \cellcolor{bestgreen}\textbf{67.2} & 74.4 & 78.0 & 74.7  
  \\
      Ours (CDT retrained)      & 72.9 & 75.9 & \cellcolor{bestgreen}\textbf{74.5} & 71.6 & \cellcolor{bestgreen}\textbf{78.1} & 75.6 & 61.5 & 74.3 & 79.9 & 73.8 \\                             
      Ours (CDT + Adaptation)   & \cellcolor{bestgreen}\textbf{74.1} & 76.7 & 73.2 & \cellcolor{bestgreen}\textbf{78.0} & 76.6 & \cellcolor{bestgreen}\textbf{80.0} & 62.5 &                     
  \cellcolor{bestgreen}\textbf{76.3} & \cellcolor{bestgreen}\textbf{81.0} & \cellcolor{bestgreen}\textbf{75.4} \\                                                                                
      \bottomrule                                                                                                                                                                                
      \end{tabular}                                                                                                                                                                              
      }           
      \caption{Temporal Adaptation Study — 4-Dim (Qwen2.5-7B-Instruct).}
      \label{tab:qwen_adapt_4dim}
  \end{table*}

\FloatBarrier

    \FloatBarrier
\section{Prompt Templates}
\label{app:prompts}

This appendix collects all prompt templates used across CDT construction (\S\ref{sec:construction}), adaptation (\S\ref{sec:adaptation}), and inference. Placeholders enclosed in braces (e.g., \texttt{\{group\}}) are filled at runtime.


\begin{figure}[h!]
\begin{tcolorbox}[colback=gray!5, colframe=gray!50, fontupper=\small\ttfamily, left=4pt, right=4pt, top=4pt, bottom=4pt]
\# Scene-Action Pairs\\
\{action\_scene\_context\}\\[4pt]
\# Established Statements\\
\{established\_statement\_verbalized\}\\[4pt]
\# Already Proposed Common Points\\
\{gate\_path\_verbalized\}\\[4pt]
\# Task\\[2pt]
Your task is to build the grounding logic for an AI system to understand the behavior of \{group\_name\} (Current topic: "\{goal\_topic\}"), assert the AI system has no prior knowledge of \{group\_name\}.\\
To do this, please propose hypotheses for the general behavior logic of \{group\_name\} based on the given action-scene pairs, complete the task step by step:\\[2pt]
1. What's the main feature of \{group\_name\}'s behavior (Focus on the current topic: "\{goal\_topic\}") shown in the given scene-action pairs, \textbf{other than the already established statements}?\\[2pt]
2. Summarize \{k\} potential common points (grounding statements) of the actions taken by \{group\_name\} in the given scenes about the focused topic: "\{goal\_topic\}", \textbf{which is other than the already established statements}.\\
- The grounding statements should be general, avoiding too specific action descriptions.\\
- Consider the grounding statements in a general way.\\
- The grounding statements should be concise, informative, and general sentences.\\
- Never be assertive! Always make objective description of the character rather than making assertive causal relations.\\
- Keep each statement decision-relevant: it should explain *why this subset of actions* happens, not just a broad institutional slogan.\\[2pt]
3. Summarize \{k\} potential common points of the given scenes that trigger each behavior, \textbf{which should be different from already proposed common points.}\\
- The question should be simple, not ambiguous, and specific to a subset of scenes rather than always applicable.\\
- Focus on the \textbf{next action} when asking! Don't ask whether certain event is involved, instead ask whether the scene might trigger potential behavior for \{group\_name\}'s \textbf{next action}.\\
- Directly include "\{group\_name\}'s next action" in the question!\\
- Make each question selective (rough target 20\%-70\% scene coverage), avoid near-universal questions.\\
- Use observable scene conditions (governance constraints, funding pressure, political climate, student demand, reputational stakes).\\[2pt]
4. Output the hypothesized scene-action triggers in the following format:\\
\texttt{action\_hypotheses = [~]  \# A list of grounding statements (strings)}\\
\texttt{scene\_check\_hypotheses = [~]  \# A list of syntactically complete questions to check the given scene (always mentioning \{group\_name\})}
\end{tcolorbox}
\caption{Hypothesis generation prompt (Step~2 of CDT construction). Given clustered scene--action pairs and existing statements, the LLM proposes $k$ behavioral statements and $k$ corresponding gate questions via chain-of-thought reasoning.}
\label{fig:prompt-hypo-gen}
\end{figure}

\begin{figure}[h]
\begin{tcolorbox}[colback=gray!5, colframe=gray!50, fontupper=\small\ttfamily, left=4pt, right=4pt, top=4pt, bottom=4pt]
\# Task: Summarize \& Compress Scene--Action Hypothesis Pairs\\[2pt]
You are given a list of \{n\_input\} paired hypotheses. Each pair contains:\\
- "scene\_check\_hypothesis": a question about \{group\_name\}'s next action\\
- "action\_hypothesis": a general behavioral grounding statement about \{group\_name\}\\[4pt]
Input pairs:\\
\{paired\_hypotheses\_json\}\\[4pt]
\#\# Goal\\
Produce a rewritten, deduplicated, and compressed set of pairs that capture the \textbf{most important} and \textbf{most general} behavioral grounding logic for \{group\_name\}.\\[2pt]
You should output between \{n\_target\} and \{n\_upper\} pairs --- use your judgment to keep as many \textbf{meaningfully distinct} pairs as needed, but merge or drop redundant ones.\\[2pt]
Rewriting is allowed and encouraged to increase: generality, coverage across different subsets of scenes, clarity, non-assertiveness.\\[4pt]
\#\# Selection Principles (prioritized)\\
1. \textbf{Coverage}: The pairs should collectively cover the widest range of distinct behavioral patterns and distinct scene triggers.\\
2. \textbf{Centrality}: Prefer pairs that reflect recurring or core behaviors across many scene-action pairs.\\
3. \textbf{Specificity without overfitting}: Keep statements general; only keep a specific skill/ability if it appears repeatedly and broadly.\\
4. \textbf{Non-redundancy}: Each pair must represent a meaningfully different behavior/trigger from the others.\\
5. \textbf{Pair coherence}: The scene\_check\_hypothesis must plausibly test for the corresponding action\_hypothesis.\\
6. \textbf{Gate selectivity}: Prefer scene\_check\_hypothesis that are neither almost always true nor almost always false.\\[4pt]
\#\# Output Format (JSON only)\\
\{"pairs": [\{"scene\_check\_hypothesis": "...", "action\_hypothesis": "..."\}]\}
\end{tcolorbox}
\caption{Hypothesis summarization and deduplication prompt (Step~3). Merges overlapping hypotheses from multiple clusters into a compact, non-redundant set.}
\label{fig:prompt-hypo-summarize}
\end{figure}

\begin{figure}[h]
\begin{tcolorbox}[colback=gray!5, colframe=gray!50, fontupper=\small\ttfamily, left=4pt, right=4pt, top=4pt, bottom=4pt]
Group: \{$g$\}\\[4pt]
Action: \{$d_i$\}\\[4pt]
Statement: \{$\hat{s}$\}\\[4pt]
Question: Is the action consistent with the behavioral pattern described in the statement?\\[2pt]
yes: the action follows or reflects the pattern described in the statement.\\
no: the action is unrelated to or contradicts the pattern described in the statement.\\[2pt]
Directly answer only yes/no.
\end{tcolorbox}
\caption{Ungated validation prompt (Step~4, Stage~1). The discriminator LLM judges whether a single action is consistent with a candidate behavioral statement without any gate conditioning.}
\label{fig:prompt-coverage}
\end{figure}

\begin{figure}[h]
\begin{tcolorbox}[colback=gray!5, colframe=gray!50, fontupper=\small\ttfamily, left=4pt, right=4pt, top=4pt, bottom=4pt]
Scene: \{$c_i$\}\\[4pt]
Question: \{$\hat{q}$\}\\[4pt]
Answer yes or no based on available evidence. Answer unknown only when the scene is completely unrelated to the question.
\end{tcolorbox}
\caption{Gate evaluation prompt (Step~4, Stage~2). Determines whether a scene satisfies a candidate gate question. Used during both construction and inference-time tree traversal.}
\label{fig:prompt-gate-check}
\end{figure}

\begin{figure}[h]
\begin{tcolorbox}[colback=gray!5, colframe=gray!50, fontupper=\small\ttfamily, left=4pt, right=4pt, top=4pt, bottom=4pt]
I have generated \{$C$\} candidate Codified Decision Trees (CDTs) intended to model the behavior of the group "\{group\_name\}".\\
Please evaluate them and select the best one based on:\\
1. Coherence and logic of the decision flow.\\
2. Generalized understanding of the group's behavior (avoiding overfitting to specific trivial details).\\
3. Clarity and meaningfulness of the gates (questions) and statements (behaviors).\\[4pt]
Here are the candidates:\\[2pt]
\{verbalized\_candidates\}\\[4pt]
Task:\\
1. Analyze the strengths and weaknesses of each candidate briefly.\\
2. Select the single best candidate.\\
3. Output your choice in the following JSON format:\\
\{"best\_candidate\_index": <1-based index>, "reasoning": "<your reasoning>"\}
\end{tcolorbox}
\caption{Multi-candidate selection prompt (Step~5). Each of $R_{\mathrm{sel}}$ voting rounds presents the candidates in a random order; the candidate with the most votes is selected.}
\label{fig:prompt-candidate-select}
\end{figure}

\FloatBarrier

\begin{figure}[h]
\begin{tcolorbox}[colback=gray!5, colframe=gray!50, fontupper=\small\ttfamily, left=4pt, right=4pt, top=4pt, bottom=4pt]
Group: \{$g$\}\\[4pt]
Action: \{$d_i$\}\\[4pt]
Classify the relationship between the action and EACH statement below.\\[2pt]
For each statement, answer:\\
- supports: the action follows, reflects, or is consistent with the pattern.\\
- irrelevant: the action is unrelated to the statement.\\
- contradicts: the action conflicts with the pattern.\\[4pt]
Statements:\\
{[1] \{$s_1$\} [2] \{$s_2$\} \ldots}\\[4pt]
Output JSON only:\\
{["supports or irrelevant or contradicts", ...]}\\
Return a JSON array with exactly one label per statement, in the same order.
\end{tcolorbox}
\caption{Batch NLI classification prompt (Phase~1). Classifies the relationship between one action and all statements at a node in a single call, enabling efficient incremental update of the relation matrix $\mathbf{M}_n$.}
\label{fig:prompt-nli-batch}
\end{figure}

\begin{figure}[h]
\begin{tcolorbox}[colback=gray!5, colframe=gray!50, fontupper=\small\ttfamily, left=4pt, right=4pt, top=4pt, bottom=4pt]
You are analyzing behavioral patterns of \{$g$\}.\\[4pt]
\#\# Statement being demoted\\
"\{$s_j$\}"\\[2pt]
This statement has precision \{precision\} at the current node --- it holds for some events but not others. We need to find a scene condition that separates the supporting events from the contradicting events, so the statement can be moved to a more specific subtree.\\[4pt]
\#\# Supporting events (action consistent with the statement):\\
\{sup\_events\}\\[4pt]
\#\# Contradicting events (action conflicts with the statement):\\
\{con\_events\}\\[4pt]
\#\# Task\\
Generate 3 candidate yes/no gate questions about the scene context that would separate the supporting events from the contradicting/irrelevant ones. Each question should:\\
- Be about observable scene conditions (not about the action itself)\\
- Be specific enough to distinguish this subset of events\\
- Always mention "\{$g$\}" or reference their situation\\
- Be answerable with yes/no from the scene context alone\\
- Each candidate should take a different angle\\[4pt]
Output as JSON: ["question 1", "question 2", "question 3"]
\end{tcolorbox}
\caption{Gate hypothesis generation prompt (Phase~2, demotion). When a statement's precision falls below $\tau_{\mathrm{keep}}$, this prompt generates candidate gate questions to separate supporting from contradicting observations.}
\label{fig:prompt-gate-hypo}
\end{figure}

\begin{figure}[h]
\begin{tcolorbox}[colback=gray!5, colframe=gray!50, fontupper=\small\ttfamily, left=4pt, right=4pt, top=4pt, bottom=4pt]
You are analyzing behavioral patterns of \{$g$\}.\\[4pt]
\#\# Statement\\
"\{$s_j$\}"\\[4pt]
\#\# Gate question\\
"\{$q_k$\}"\\[4pt]
\#\# Task\\
Does this gate question provide a meaningful scene condition under which the statement would be specifically relevant? In other words, is the statement a natural behavioral pattern to expect when the gate condition is true?\\[4pt]
Answer only: yes or no.
\end{tcolorbox}
\caption{Gate--statement semantic compatibility check (Phase~2, demotion). Verifies that a candidate gate question is semantically coherent with the statement being demoted to a child subtree.}
\label{fig:prompt-gate-semantic}
\end{figure}

\begin{figure}[h]
\begin{tcolorbox}[colback=gray!5, colframe=gray!50, fontupper=\small\ttfamily, left=4pt, right=4pt, top=4pt, bottom=4pt]
You are analyzing behavioral patterns of \{$g$\}.\\[4pt]
\#\# CDT path from Root to this node:\\
\{path\_text\}\\[4pt]
\#\# Uncovered events at this node:\\
These events are not supported by any existing statement at this node.\\
\{uncovered\_events\}\\[4pt]
\#\# Existing statements at this node (for reference, do not duplicate):\\
\{existing\_statements\}\\[4pt]
\#\# Task\\
Generate new behavioral statements that capture the patterns in the uncovered events.\\
- Each statement should be one sentence, specific to the topic indicated by the gate path.\\
- Do NOT duplicate or rephrase existing statements.\\
- Focus on the behavioral pattern, not specific actions.\\[4pt]
Output as JSON: \{"statements": ["statement 1", "statement 2"]\}
\end{tcolorbox}
\caption{New statement generation prompt (Phase~3). Proposes behavioral statements for observations not covered by any existing statement at the current node, conditioned on the gate path from root.}
\label{fig:prompt-add-stmt}
\end{figure}

\FloatBarrier

\begin{figure}[h]
\begin{tcolorbox}[colback=gray!5, colframe=gray!50, fontupper=\small\ttfamily, left=4pt, right=4pt, top=4pt, bottom=4pt]
\# Background Knowledge\\
\{background\}\\[4pt]
\# Context\\
\{$c^*$\}\\[4pt]
\# Question\\
\{question\}\\[4pt]
Predict the specific action taken by \{$g$\}. State the concrete decision, not the motivation or background. Answer in one sentence.
\end{tcolorbox}
\caption{CDT inference prompt.}
\label{fig:prompt-infer-bg}
\end{figure}

\FloatBarrier

\begin{figure}[h]
\begin{tcolorbox}[colback=gray!5, colframe=gray!50, fontupper=\small\ttfamily, left=4pt, right=4pt, top=4pt, bottom=4pt]
\textbf{Context:} \{$c^*$\} \\[4pt]
\textbf{Premise:} \{$d^*$\} \\
\textbf{Hypothesis:} \{$\hat{d}$\} \\[4pt]
Determine the relationship between the premise and hypothesis. \\[2pt]
- ``entails'': The hypothesis can be inferred from the premise. They describe the same action or event. \\
- ``neutral'': The hypothesis is neither supported nor contradicted by the premise. \\
- ``contradicts'': The hypothesis is incompatible with the premise.
\end{tcolorbox}
\caption{NLI-based evaluation prompt for consistency-level evaluation.}
\label{fig:prompt-nli-eval}
\end{figure}

\begin{figure}[h]
\begin{tcolorbox}[colback=gray!5, colframe=gray!50, fontupper=\small\ttfamily, left=4pt, right=4pt, top=4pt, bottom=4pt]
\# Context\\
\{$c^*$\}\\[2pt]
\# Your Response: \{prediction\}\\
\# Ground Truth: \{reference\}\\[4pt]
Compare the action of \{$g$\} in the response against the ground truth. Focus on whether the \textbf{strategic character} of the actions aligns, not whether the specific actions are identical.\\[4pt]
\textbf{Initiative} --- Whether both actions are driven by the same type of trigger.\\
\hspace*{1em}Proactive: the entity initiates a new move on its own accord.\\
\hspace*{1em}Reactive: the entity responds to external events or pressures.\\
\hspace*{1em}Match if both share the same trigger type; mismatch otherwise.\\[4pt]
Output: \{``initiative'': ``match'' | ``mismatch'', ``reason'': ``...''\}
\end{tcolorbox}
\caption{Evaluation prompt for initiative dimension.}
\label{fig:prompt-initiative}
\end{figure}

\begin{figure}[h]
\begin{tcolorbox}[colback=gray!5, colframe=gray!50, fontupper=\small\ttfamily, left=4pt, right=4pt, top=4pt, bottom=4pt]
\# Context\\
\{$c^*$\}\\[2pt]
\# Your Response: \{prediction\}\\
\# Ground Truth: \{reference\}\\[4pt]
Compare the action of \{$g$\} in the response against the ground truth.\\[4pt]
\textbf{Scope} --- Whether both actions are directed at the same domain.\\
\hspace*{1em}Internal: directed inward (restructuring, reform, resource reallocation).\\
\hspace*{1em}External: directed outward (market expansion, product launch, partnership).\\
\hspace*{1em}Match if both target the same domain; mismatch otherwise.\\[4pt]
Output: \{``scope'': ``match'' | ``mismatch'', ``reason'': ``...''\}
\end{tcolorbox}
\caption{Evaluation prompt for scope dimension.}
\label{fig:prompt-scope}
\end{figure}

\begin{figure}[h]
\begin{tcolorbox}[colback=gray!5, colframe=gray!50, fontupper=\small\ttfamily, left=4pt, right=4pt, top=4pt, bottom=4pt]
\# Context\\
\{$c^*$\}\\[2pt]
\# Your Response: \{prediction\}\\
\# Ground Truth: \{reference\}\\[4pt]
Compare the action of \{$g$\} in the response against the ground truth.\\[4pt]
\textbf{Magnitude} --- Whether both actions represent a similar scale of change.\\
\hspace*{1em}Incremental: minor adjustment or refinement.\\
\hspace*{1em}Moderate: notable but bounded change.\\
\hspace*{1em}Transformative: fundamental strategic shift.\\
\hspace*{1em}Match if both are at the same or adjacent levels; mismatch if non-adjacent.\\[4pt]
Output: \{``magnitude'': ``match'' | ``mismatch'', ``reason'': ``...''\}
\end{tcolorbox}
\caption{Evaluation prompt for magnitude dimension.}
\label{fig:prompt-magnitude}
\end{figure}

\begin{figure}[h]
\begin{tcolorbox}[colback=gray!5, colframe=gray!50, fontupper=\small\ttfamily, left=4pt, right=4pt, top=4pt, bottom=4pt]
\# Context\\
\{$c^*$\}\\[2pt]
\# Your Response: \{prediction\}\\
\# Ground Truth: \{reference\}\\[4pt]
Compare the action of \{$g$\} in the response against the ground truth.\\[4pt]
\textbf{Horizon} --- Whether both actions operate on the same time horizon.\\
\hspace*{1em}Exploitative: short-term optimization, immediate response.\\
\hspace*{1em}Explorative: long-term investment, building new capabilities.\\
\hspace*{1em}Match if both serve the same timeframe; mismatch otherwise.\\
\hspace*{1em}Note: forward-looking language does not make an action explorative; judge by what it actually accomplishes.\\[4pt]
Output: \{``horizon'': ``match'' | ``mismatch'', ``reason'': ``...''\}
\end{tcolorbox}
\caption{Evaluation prompt for horizon dimension.}
\label{fig:prompt-horizon}
\end{figure}

\FloatBarrier

\begin{figure}[h]
\begin{tcolorbox}[colback=gray!5, colframe=gray!50, fontupper=\small\ttfamily, left=4pt, right=4pt, top=4pt, bottom=4pt]
\# Context\\
\{$c^*$\}\\[4pt]
\# Question\\
\{question\}\\[4pt]
Predict the specific action taken by \{$g$\}. State the concrete decision, not the motivation or background. Answer in one sentence.
\end{tcolorbox}
\caption{Vanilla inference prompt (no background knowledge).}
\label{fig:prompt-infer-base}
\end{figure}

\begin{figure}[h]
\begin{tcolorbox}[colback=gray!5, colframe=gray!50, fontupper=\small\ttfamily, left=4pt, right=4pt, top=4pt, bottom=4pt]
\# Background Knowledge\\
\{background\}\\[4pt]
\# Scene\\
\{$c^*$\}\\[4pt]
\# Question\\
\{question\} Answer a concise narration in one sentence.
\end{tcolorbox}
\caption{Human-profile inference prompt.}
\label{fig:prompt-vanilla}
\end{figure}

\begin{figure}[h]
\begin{tcolorbox}[colback=gray!5, colframe=gray!50, fontupper=\small\ttfamily, left=4pt, right=4pt, top=4pt, bottom=4pt]
\# Task\\
Please provide a 1000-word, narrative-style character profile for \{character\}.\\
The profile should read like a cohesive introduction, weaving together the character's background, personality traits and core motivations, notable attributes, relationships, key experiences, major decisions or actions, and character arc or development.\\
The profile should be written in a concise yet informative style, similar to what one might find in a comprehensive character guide. Focus on the most crucial information that gives readers a clear understanding of the character's significance.\\
The profile should be based on either your existing knowledge of the character or the provided information, without fabricating or inferring any inaccurate or uncertain details.\\[4pt]
\# Scene-Action Pairs\\
\{block\}\\[4pt]
Now, based on the given scene-action pairs, please generate the character profile, starting with ===Profile===.
\end{tcolorbox}
\caption{Summarization-based profile extraction prompt.}
\label{fig:prompt-eta-extract}
\end{figure}

\begin{figure}[h]
\begin{tcolorbox}[colback=gray!5, colframe=gray!50, fontupper=\small\ttfamily, left=4pt, right=4pt, top=4pt, bottom=4pt]
\# Main Profile\\
\{main\_profile\}\\[4pt]
\# New Summarized Profile (From New Episodes)\\
\{summarized\_profile\}\\[4pt]
Directly update the main profile based on the new summarized profile, keep its length in around 1000 words.
\end{tcolorbox}
\caption{Summarization-based profile aggregation prompt.}
\label{fig:prompt-eta-aggregate}
\end{figure}

\begin{figure}[h]
\begin{tcolorbox}[colback=gray!5, colframe=gray!50, fontupper=\small\ttfamily, left=4pt, right=4pt, top=4pt, bottom=4pt]
\# In-Context Examples\\
The following are past scene-action pairs for \{$g$\}:\\[2pt]
\{examples\}\\[4pt]
\# Context\\
\{$c^*$\}\\[4pt]
\# Question\\
\{question\}\\[4pt]
Predict the specific action taken by \{$g$\}. State the concrete decision, not the motivation or background. Answer in one sentence.
\end{tcolorbox}
\caption{RAG inference prompt.}
\label{fig:prompt-ricl}
\end{figure}

\end{document}